%% file: iclr2027_conference.tex
\documentclass{article} 
\usepackage{iclr2027_conference,times}

\input{math_commands.tex}

\usepackage{xspace}
\usepackage[colorlinks]{hyperref}
\usepackage{url}
\usepackage{graphicx}
\usepackage{tcolorbox}
\usepackage{algpseudocode}
\tcbuselibrary{breakable}

\usepackage{algorithm}
\usepackage{enumitem}
\usepackage{cleveref}

\input{preamble}

\usepackage{bm}
\usepackage{tcolorbox}
\tcbuselibrary{breakable}

\def\onedot{\xspace}

\def\eg{\emph{e.g}\onedot} 
\def\ie{\emph{i.e}\onedot}

\usepackage{pifont}
\usepackage{etoc}
\usepackage{booktabs} 

\def\name{HARMONY\xspace}

\title{\name: Hierarchical Agentic Reasoning for MONocular Image-to-Scene Synthesis}

\author{
~~~~~~~~~~~~~~~~~Shufan Sun$^{*}$ \quad
Chen Wang$^{*}$ \quad
Enxin Song \quad
Jiatao Gu \quad
Lingjie Liu \\
\\
~~~~~~~~~~~~~~~~~~~~~~~~~~~~~~~~~~~~~~~~~~~~~~~~~~~~~~University of Pennsylvania\\
~~~~~~~~~~~~~~~~~~~~~~~~~~~~~~~~~~~~\url{https://cwchenwang.github.io/harmony}\\
~~~~~~~~~~~~~~~~~~~~~~~~~~~~~~~~~~~~~~~~~~~~~~~~~~
\vspace{-4ex}
}

\footnotetext[1]{Equal contribution.}

\iclrfinalcopy 
\begin{document}

\maketitle

\vspace{-15pt}

\begin{figure}[h]
\begin{center}
\includegraphics[width=\linewidth]{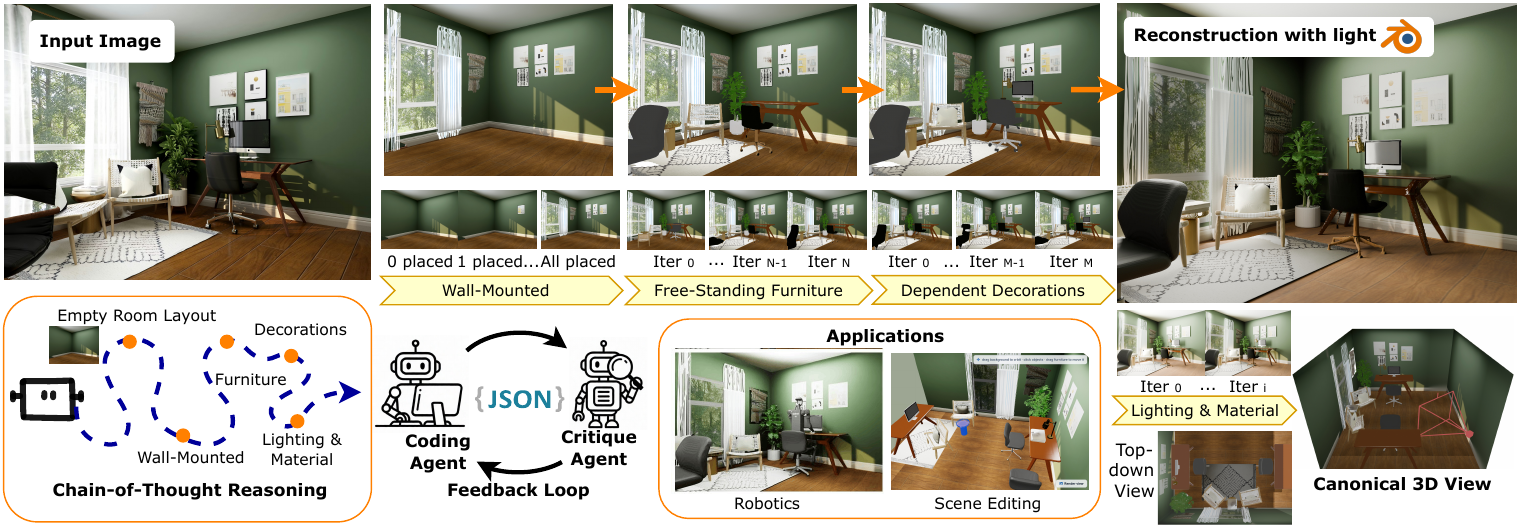}
\end{center}
\caption{Given a single input image, we reconstruct individual objects and use a VLM to reason object placements and orientations, yielding a high-quality compositional 3D scene. 
  }
  \label{fig:teaser}
\end{figure}

\input{sections/00_abstract}

\input{sections/01_introduction}
\input{sections/02_related_work}
\input{sections/03_method}

\input{sections/04_experiments}
\input{sections/05_conclusion}

\bibliography{iclr2027_conference}
\bibliographystyle{iclr2027_conference}

\appendix
\input{sections/X_suppl}

\end{document}

%% file: math_commands.tex
\usepackage{amsmath,amsfonts,bm}

\def\eqref#1{equation~\ref{#1}}

\def\1{\bm{1}}

\DeclareMathAlphabet{\mathsfit}{\encodingdefault}{\sfdefault}{m}{sl}
\SetMathAlphabet{\mathsfit}{bold}{\encodingdefault}{\sfdefault}{bx}{n}



%% file: preamble.tex
\usepackage[dvipsnames]{xcolor}
\usepackage{booktabs}
\usepackage{multirow}
\usepackage{listings}
\usepackage{cuted}
\usepackage{caption}
\usepackage{xspace}
\usepackage{float}
\newcommand{\topic}[1]
{
\noindent\textbf{#1}
}



%% file: sections/00_abstract.tex
\begin{abstract}
Compositional 3D scene reconstruction has recently been explored from two directions: agentic reasoning that provides semantic understanding of spatial relationships but lacks precise alignment with input images; and visual geometry foundation models that predict dense point maps from input images but the reconstruction quality is limited. Therefore, recovering a complete 3D scene from a single monocular image with accurate inter-object relationships and high-fidelity reconstruction quality remains challenging. In this paper, we present \name, a hierarchical chain-of-thought framework that leverages both agentic reasoning and visual geometry foundation. 
Given an image of an indoor scene, starting from an empty 3D floorplan, \name first calibrates the camera against the reference image to establish a semantically-grounded spatial frame, then uses agentic VLM reasoning to recover the 3D room layout and an initial placement order. 
It then places the objects in a hierarchical order, from wall-mounted elements, free-standing furniture, to dependent decorations on top of furniture.
We also use depth-first traversal for furniture so each placement conditions on previously resolved structure and a reflective feedback loop to avoid error accumulation. 
After each object placement by VLM, we use the point cloud estimations to perform geometry-based refinement so that the rendered image aligns better with the input.
\name can produce 3D scenes that are semantically consistent and perceptually aligned with the reference image, extending single-image compositional reconstruction to complex indoor scene images.
Experiments on synthetic and real-world images demonstrate that HARMONY outperforms the evaluated reconstruction baselines, while qualitative comparisons with GPT-6 Astra suggest more faithful object arrangements and better preservation of scene details. 
\end{abstract}

%% file: sections/01_introduction.tex
\vspace{-15pt}
\section{Introduction}
\label{sec:intro}

Reconstructing a compositional 3D scene from a single image is a long-standing problem in computer graphics and vision, with applications spanning AR/VR content creation, embodied AI, robotic navigation, and interactive scene editing. 
However, recovering a 3D scene from a single image is inherently ill-posed: a single view captures only partial geometry and typically exhibits heavy inter-object occlusions. Beyond geometry, producing a plausible scene layout is also challenging, as 2D images provide no explicit cues about exact object scales and spatial relationships.

Prior work on compositional 3D scene reconstruction has largely built upon image-to-3D generation and visual geometry foundation models. Given an input image, these methods~\citep{3dregen,hiscene,Imaginarium,yao2025cast} segment and reconstruct each object separately, then assemble them into a shared coordinate system by aligning perception signals such as estimated depth and point clouds. Relying solely on low-level perceptual cues without explicit reasoning over inter-object relationships, they often struggle with occluded and small objects, producing errors in object pose and relative placement.
Another line of work leverages vision–language models (VLMs) for spatial reasoning over object relationships~\citep{scenesmith2026,viga,sage, yao2025cast}, modeling how each object interacts with the floorplan and surrounding objects rather than how it projects into a camera.
Text-conditioned methods~\citep{scenesmith2026} exploit spatial reasoning to plan a hierarchical placement order from wall-aligned items to furniture in order to recover relations such as ``against the wall" or ``on top of".
However, because they operate purely in language and ground the scene through asset retrieval, these pipelines are restricted to scenes composed of simple objects and template-like arragments. 
Image-conditioned VLM methods~\citep{sage,viga,holodeck, holodeck2} restore this grounding and achieve strong high-level semantic alignment with the observed scene, but inherit the VLM's well-known weakness of being unable to perform precise visual–geometric reasoning, yielding inaccurate object placements and noticeable visual mismatch with the input.

In this paper, we propose \name, a hierarchical VLM-guided framework for high-quality single-image to 3D scene reconstruction that leverages the strengths of both agentic reasoning and visual geometry-grounded signals. 
We first use the semantic and spatial understanding of VLM to reason and plan a 3D scene that highly aligns with the input image.
Specifically, rather than placing all objects at once, we decompose the problem into a structured reasoning process of multiple stages. The VLM first grounds itself spatially in the scene by identifying corners, walls, and viewpoint to establish a fixed frame of reference. It then reasons through object placement in a hierarchical order: wall-mounted items, free-standing furniture, and finally decorations that rest on top of the furniture.
We also use depth-first traversal to place the furniture: at the moment any object is placed, every previously placed object lies behind it from the camera's perspective. Therefore, each new candidate appears in a clean, unoccluded view of the partial scene, allowing the VLM to reason about its orientation and pairing without any interference from clutter.
After the VLM obtains a good plan of the 3D scene, we leverage visual geometry signals as a refinement tool for more fine-grained positioning, including both image-space alignment and depth-space alignment. Our design avoids the weaknesses of previous methods that rely only on point clouds, which cannot handle occlusions or small objects.
Also, after each placement stage, we introduce a reflective feedback loop that uses VLM to compare the rendering of the partial scene against the input image and identify correspondence issues such as missing items, incorrect pairings, or wrong orderings, etc. This allows us to correct and prevent errors from propagating to the next stages. 
We evaluate our method on both synthetic and real-world indoor input images, and the results demonstrate that we are able to reconstruct the 3D scene in a high-quality compositional manner.

In summary, our contributions can be summarized as the following:
\begin{itemize}
    \item Given a single image of an indoor scene, we propose a hierarchical chain-of-thought framework to reconstruct a compositional 3D scene. We frame this problem as a structured reasoning process and use VLM to ground the scene spatially and place objects in multiple stages.
    \item \name marries the complementary strengths of VLMs and visual geometry-grounded models. We first use a VLM to reason about spatial and semantic relationships, \ie, what an object leans against, sits on, faces, or pairs with. Then we adjust the precise position based on the estimated point clouds.
    \item \name achieves state-of-the-art performance in single-image compositional 3D scene reconstruction on both synthetic and challenging real-world inputs, with qualitative comparisons against GPT-6 Astra indicating more faithful object arrangements and better preservation of scene details.
\end{itemize}

%% file: sections/02_related_work.tex
\section{Related Work}

\topic{Image-to-3D Object Reconstruction}
Given an image of a 3D object, image-to-3D reconstruction outputs both the 3D geometry and textures. The current dominant paradigm is based on 3D native diffusion that trains a diffusion model directly on 3D representations.
3DShape2VecSet~\citep{zhang20233dshape2vecset} pioneers this line of research, encoding shapes
into latents with cross-attention that can be decoded to occupancy fields.
CLAY~\citep{zhang2024clay} scales latent-set diffusion to billion-scale parameters on large-scale 3D data, and TRELLIS~\citep{xiang2024structured} unifies geometry and appearance in a structured latent that decodes to multiple representations, including 3D Gaussians, radiance fields, and meshes. 
Recent works have further pushed scale, fidelity, and material expressiveness based on the latent diffusion transformer.
The Hunyuan3D~2 series~\citep{zhao2025hunyuan3d2}
progressively adds PBR materials and finer geometric detail; TripoSG~\citep{li2025triposg} adopts a rectified-flow transformer with larger latent capacity; and Direct3D-S2~\citep{wu2025direct3ds2} introduces sparse attention for gigascale training. TRELLIS.2~\citep{xiang2025trellis2} extends the structured-latent design with a field-free O-Voxel representation that jointly handles arbitrary topology and PBR appearance.
From a complementary angle, SAM3D~\citep{sam3dteam2025sam3d3dfy} introduces human-in-the-loop annotation for strong reconstructions on in-the-wild images with heavy occlusion. Our method integrates these advances in object reconstruction with hierarchical reasoning and geometry-grounded refinement to assemble coherent scenes with object scales, orientations, and spatial relationships aligned with the input image.


\topic{Geometry-Grounded Image-to-3D Scene Reconstruction}
Recent advances in visual geometry learning and 2D segmentation have facilitated the reconstruction of single-view input images.
A typical line of methods decomposes the problem into multiple stages, including point cloud estimation~\citep{vggt}, segmentation~\citep{kirillov2023segany,liu2023grounding,ren2024grounded}, context-aware inpainting~\citep{qwen3technicalreport,Qwen2.5-VL,Qwen2VL,Qwen-VL,Gemini2026Image}, single object reconstruction~\citep{xiang2025trellis2, zhao2025hunyuan3d2, sam3dteam2025sam3d3dfy}, and finally layout optimization.
For example, Gen3DSR~\citep{dogaru2025gen3dsr} applies a divide-and-conquer strategy that pairs holistic scene parsing with object-level generative reconstruction and ZeroScene~\citep{tang2026zeroscene} optimizes
per-object poses by jointly minimizing 3D and 2D projection losses on segmented point clouds.
CAST~\citep{yao2025cast} reasons about inter-object spatial relations through a GPT-based scene parser, employs an occlusion-aware large 3D generation model for each component, and resolves penetration and floating artifacts through SDF-based physical correction. 3D-RE-GEN~\citep{3dregen} extends this idea with explicit background reconstruction and a 4-DoF differentiable optimization that aligns reconstructed objects to the estimated ground plane. 
A complementary direction trains a single network to directly predict the whole scene. Coherent 3D Scene Diffusion~\citep{dahnert2024coherent} and MIDI~\citep{huang2025midi} jointly diffuse all objects' shapes and poses with cross-instance attention, while SceneGen~\citep{meng2025scenegen} produces all 3D assets in a single feed-forward pass without per-object optimization.
However, these pipelines lack semantic spatial understanding for resolving object orientation, so they often recover noisy facings and miss inter-object relationships, such as how a chair should face relative to a desk. 



\topic{Agentic Reasoning for 3D Scene Generation}
Advances in multimodal vision-language models (VLMs)~\citep{Qwen-VL,Qwen2.5-VL,Qwen2VL,qwen3technicalreport} have enabled reasoning about object arrangements and scene graphs based on semantics. 
Given a text description of a scene, SceneSmith~\citep{scenesmith2026}, for instance, uses a VLM to initialize a floorplan with wall dimensions, then reasons about a hierarchical placement order that captures how objects relate to one another and to the surrounding layout. 
Moreover, because text descriptions are not able to describe accurate 3D positions and orientations, these methods often rely on asset retrieval and hand-designed priors, such as canonical object orientations like the canonical facing direction of a bed in a bedroom. 

Adding a reference image to VLM generation provides a more concrete grounding signal. Holodeck and Holodeck2.0~\citep{holodeck2, yang2024holodeck} use a VLM to parse objects and emit constraint relations that drive a layout solver, while SAGE~\citep{sage} converts the image to text descriptions and only align the image semantically. 
VIGA~\citep{viga} builds a Blender agent that iteratively adjusts the reconstructed scene by rendering and comparing it with the input. However, it still only produces scenes that are semantically similar to the reference because VLM itself cannot reason precise numerical quantities.
Spatially-Contextualized VLMs~\citep{contextualized_agentic} augment VLM reasoning with explicit perception signals, \ie, point clouds produced by Fast3R~\citep{Yang_2025_Fast3R}. However, small objects such as decorations can be difficult to resolve in monocular images, leading to incomplete or noisy point-cloud estimates. Our work proposes a hierarchical reasoning pipeline and performs checking in each stage to reduce error accumulation.

%% file: sections/03_method.tex
\begin{figure*}[!t]
  \centering
  \includegraphics[width=\linewidth]{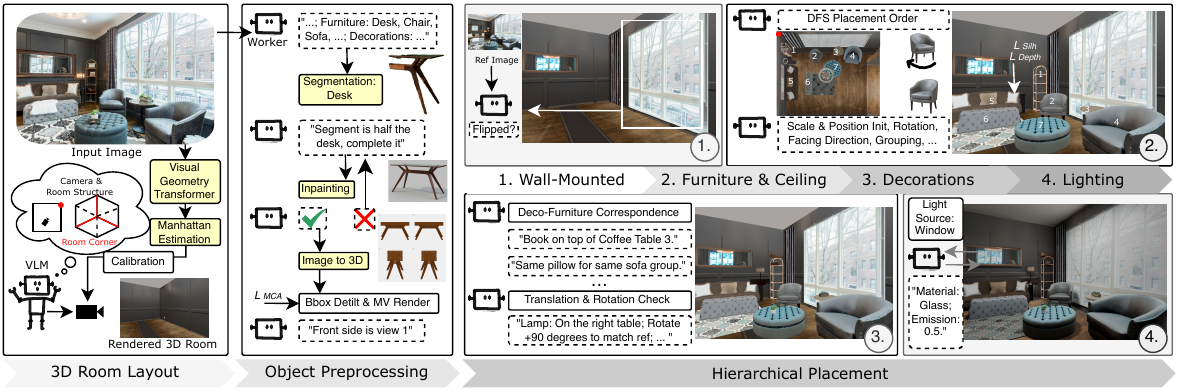}
  \vspace{-15pt}
  \caption{Overview of our pipeline. \name{} takes a monocular image as input. It first creates an empty 3D room layout and anchors the camera to the 3D scene based on the estimated point cloud and Manhattan frame. It then segments and inpaints the objects in the scene and reconstructs their 3D meshes. Next, a VLM hierarchically plans and places the objects in three stages, with point clouds used for geometric correction. Finally, \name relights the reconstructed scene using VLM-estimated material and emission properties.}
  \vspace{-15pt}
  \label{fig:pipeline}
\end{figure*}

\section{Method}
Given a monocular image of an indoor scene, our goal is to reconstruct a compositional 3D scene that faithfully recovers all objects together with their spatial relationships, such that renderings of the reconstructed scene closely match the input view. To this end, \name integrates VLM-based relational and spatial reasoning, 2D and 3D generation, and visual geometry-grounded models into a unified and scalable pipeline.

As shown in \Cref{fig:pipeline}, the backbone of \name is a hierarchical chain-of-thought framework based on VLM. Starting from an empty 3D room, we first estimate the camera pose that aligns with the perspective of the input view, anchoring its initial understanding of the scene (\Cref{subsec:3d-room-layout}). 
Next, we segment and inpaint each object, and then reconstruct them into 3D meshes (\Cref{subsec:object-process}).
Then, the VLM reasons about object placement in a hierarchical order (\Cref{subsec:recon}): first wall-mounted objects, then furniture and ceiling objects, and finally decorations. \name explicitly models spatial relationships such as what an object leans against, sits on, faces, or is paired with. Visual geometry cues are further leveraged to refine each object's scale and position.
In each stage, we also introduce a reflective feedback loop that uses the VLM to critique the rendered scene against the reference image, identifying missing items, mismatched sizes, incorrect pairings, or wrong orderings, and issues targeted corrections.
This reflective feedback loop refinement progressively reduces error accumulation and yields a scene that is both globally consistent and locally faithful to the input. 
Finally, \name uses the VLM to estimate per-object materials and the scene's emissive light sources for a physically-based render (\Cref{subsec:relit}).

\subsection{3D Room Layout and Camera Initialization}
\label{subsec:3d-room-layout}
In this stage, we aim to obtain a mesh of an empty room (i.e., walls without objects) with a camera pose that projects a layout that aligns with the reference image.
Our solution combines both semantic room understanding from the VLM and geometric corner detection from VGGT.

\subsubsection{VLM Semantic Initialization.}
As shown in the leftmost column of \Cref{fig:pipeline}, we first use the VLM to infer approximate room dimensions from semantic cues in the reference image (e.g., $(3,4,3) \mathrm{m}$ for a bedroom). These estimates serve as a scale prior for initializing the room geometry, consisting of walls, a floor, and a ceiling.
The VLM then identifies the deepest visible room corner as a spatial anchor (or the farthest wall endpoint if only walls are visible) from the reference image. 
We associate the identified anchor with the corresponding vertical edge of the canonical room mesh, whose floor and ceiling endpoints are denoted by $\widehat{\bm X}_f$ and $\widehat{\bm X}_c$, with estimated room height
$H_{\rm room}=|\widehat{\bm X}_c-\widehat{\bm X}_f|_2$.
The VLM further labels each visible wall relative to this anchor, establishing which surface regions in the canonical mesh correspond to which image walls. 
After this stage, we have an empty room box with semantically-aligned corners and planes. 

\subsubsection{VGGT Geometric Refinement.}
To anchor the same corner and its adjoining floor–wall boundaries in the VGGT reconstruction, we estimate a Manhattan frame from the predicted point cloud and VGGT camera, with extrinsics $(R_0,\mathbf t_0)$ and intrinsics $K$ (focal length $f_x$), using SVD-based clustering of surface normals, yielding three mutually orthogonal axes $\mathbf a_w,\mathbf a_v,\mathbf a_d\in\mathbb R^3$ (corresponding to the width, vertical, and depth directions), with the width–depth assignment resolved in the calibration step below.
The room's six bounding planes are located along these axes by a per-axis histogram fit and normal alignment with the surface normals.
Within this Manhattan frame we identify the deepest floor corner (farthest from the camera) $\overline{\bm X}_f$ as the intersection of the floorplane with the two wall planes meeting at it, and its ceiling counterpart directly above it, $\overline{\bm X}_c$, both in VGGT coordinate space. The Manhattan frame also identifies the two floor-wall edges extending from the anchor corner along $\mathbf a_w,\mathbf a_d$, corresponding to the width- and depth-facing walls.

\subsubsection{Camera Calibration.}
We use the VGGT camera and Manhattan frame obtained in the previous step to solve in closed form for a single similarity transform (rotation $R_{\rm align}$, uniform scale $s$, translation $\mathbf t$) that re-expresses this pose in the canonical frame, with $\overline{\bm X}_f,\overline{\bm X}_c$ aligned with the corresponding floor and ceiling endpoints of a vertical room edge. Applying this transform to the VGGT camera itself then gives its pose in the canonical frame.

\textbf{Rotation.} We first orient the canonical axes directly from the Manhattan frame: $\mathbf a_v$ is oriented upward; between $\mathbf a_w,\mathbf a_d$, whichever has the larger-magnitude dot product with the camera's forward direction $R_0^\top\mathbf e_z$ is assigned to the canonical depth axis (oriented so the camera looks toward the back wall), and the remaining axis to canonical width, with its sign fixed so that: $
R_{\rm align} =
\big[\mathbf a_w^\top;\ \mathbf a_v^\top;\ \mathbf a_d^\top\big]
$ is a proper rotation (the determinant of $R_{\rm align}$ is +1).

\textbf{Scale.} We then recover metric scale directly from the anchor edge, using the room's known height $H_{\rm room}$ as the sole external metric reference: $
s = H_{\rm room}/\lVert \overline{\bm X}_c - \overline{\bm X}_f\rVert.$.

\textbf{Translation.} We solve in closed form for the translation $\mathbf t$ that places the floor anchor exactly on its corresponding canonical wall corner $\widehat{\bm X}_f$: $
\mathbf t = \widehat{\bm X}_f - s R_{\rm align}\overline{\bm X}_f,$ giving the similarity map $\bm X \mapsto s R_{\rm align}\bm X + \mathbf t$ from VGGT space into the canonical room frame.

\textbf{Camera pose.} The camera rotation and center can thus be solved using the above similarity transform: $
R = R_0 R_{\rm align}^\top, \: \mathbf c = s R_{\rm align}\mathbf c_0 + \mathbf t,
$ where $\mathbf c_0 = -R_0^\top\mathbf t_0$ is the raw VGGT camera center.

\subsection{Object Segmentation and Reconstruction}
\label{subsec:object-process}

For compositional reconstruction, we detect and segment each object in the image, inpaint occluded ones and finally reconstruct them into 3D meshes.

\topic{Object Detection.}
We detect and segment objects hierarchically, processing one level at a time:
wall-mounted items (e.g., paintings, windows), free-standing furniture
(ground-mounted objects like desks and ceiling-mounted objects like chandeliers), and decorations that rest on furniture. At each level, the VLM parses the reference image to list the objects of that category with their per-instance counts; we pass this list to open-vocabulary detection~\citep{wang2025locateanything} for bounding boxes and then to a segmentation model~\citep{ravi2024sam2} for masks. Afterwards, we also filter duplicate and spurious detections and attach each decoration to its supporting furniture.

\topic{Object Inpainting.}
For each detected object, the VLM produces a detailed description conditioned on the surrounding scene context, which is passed together with the cropped object region to an image-editing model~\citep{Gemini2026Image} to inpaint the occluded region. A half-occluded table, for example, is described as such by the VLM, so the model can generate a complete table compatible with the scene. The VLM then inspects the inpainted result for consistency with the reference object in terms of object type, completeness, and shape alignment. If the output does not match, it regenerates with additional material and color hints.

\topic{Mesh Canonicalization and Orientation Labeling.}
After obtaining the complete image for each object, we reconstruct its 3D mesh using an image-to-3D model~\citep{zhao2025hunyuan3d2}. Since the inpainted views inherit the perspective of the input image, the resulting meshes are in non-canonical poses. We first
canonicalize each mesh by applying Principal Component Analysis (PCA) to its vertices and aligning its dominant axis with world-up. The VLM then inspects multi-view renders of the mesh and labels its facing direction, assigning a per-object canonical frame that the placement stage uses to enforce correct relative orientations between paired objects (e.g., a chair facing its companion desk).

\subsection{Hierarchical Scene Reconstruction}
\label{subsec:recon}
In this stage, the VLM reasons about the spatial relationships among objects and places them into the 3D scene in three ordered stages as in object detection: wall-mounted items, free-standing objects and decorations. We parameterize each object by its position $\mathbf p\in\mathbb{R}^3$ and a uniform scale $s$, with its yaw set by the VLM during placement. In each stage, a reflective feedback loop inspects the rendered scene and corrects errors before the pipeline proceeds.

\topic{VLM Placement Order Reasoning.}
Building on the anchor corner from the previous step, the VLM uses it as a spatial reference for performing object placements. Treating this anchor corner as the deepest point of the room, the VLM performs a depth-first traversal over the visible objects ordered by proximity to the corner: the VLM first places objects nearest the anchor along the two adjacent walls, then progressively moves outward toward the room's interior, finishing with the objects closest to the camera. 
In this way, each new object can be aligned with the existing geometry without occlusion, while its orientation and pairwise relationships are resolved within a consistent, previously established scene context.

\topic{VLM-based Object Placement.}
Once the placement order is determined, the VLM assigns an initial size to each object from prior knowledge and reasons about per-object prompts describing each object's spatial relationship to the room and other objects, \eg, ``sofa back against the left wall'', ``vase on the desk'', or ``chair facing the small coffee table''. Using each object's canonical front from the preprocessing stage (\Cref{subsec:object-process}), the VLM then sets its orientation from these relations, \eg, turning the sofa's back toward the specified wall and rotating it to face the specified neighbor. Finally, the VLM visually compares the rendered object against the reference image and applies small rotation adjustments to better match it. For decorations, the VLM identifies which previously placed piece of furniture supports each decoration and verifies that it rests on the correct one. Since the VLM's semantic reasoning can confuse furniture that shares a category label (\eg, two similar end tables), we add a consistency check to prevent misattachment, such as a vase placed on the wrong table. When the VLM flags such a case, it triggers a bounding-box check that compares the decoration's box against those of the candidate hosts and reassigns the decoration to the host whose box it overlaps most.


\topic{Visual Geometry-Grounded Refinement}
The placements produced by the VLM capture the correct semantic relations between objects but only approximate scale and location. We refine each placement in two coupled stages: the image-space silhouette fixes the object's lateral position and its scale-to-depth ratio $s/Z$ from the apparent silhouette width, while the point cloud fixes the forward distance $Z$, which in turn resolves the
metric scale $s$.

For image-space alignment, we render the object silhouette with horizontal center $u_c$ and width $w$, and align it with its segmentation mask (center $\hat u_c$, width $\hat w$). With the current object depth $Z=(\mathbf p-\mathbf c)\cdot\hat{\mathbf f}$, the horizontal offset $\Delta u=\hat u_c-u_c$ back-projects to a lateral translation $\mathbf p \leftarrow \mathbf p + \frac{\Delta u\,Z}{f_x}\,\hat{\mathbf r}$. For objects not flagged as heavily occluded, we further read the object's apparent (angular) size from the silhouette, $\alpha = \frac{\hat w}{f_x} = \frac{s\,w_{\text{obj}}}{Z}$, where $w_{\text{obj}}$ is the object's canonical width. The silhouette is inherently ambiguous between object scale and depth, so $\alpha$ determines only the ratio $s/Z$; we therefore refine the VLM's coarse size prior but defer the metric scale to the depth stage.

For depth-space alignment, we leverage the point cloud estimated by VGGT~\citep{vggt} and segment it with the per-object mask, and likewise lift the rendered scene within the rendered mask. For each object, we extract the robust front-surface means
$\mathbf c^{\text{front}}_{\text{ref}},\mathbf c^{\text{front}}_{\text{ren}}$ by iterative median with MAD outlier rejection, and correct the depth by their displacement along the view direction:
$Z' = Z + \big(\mathbf c^{\text{front}}_{\text{ref}}-\mathbf c^{\text{front}}_{\text{ren}}\big)\cdot\hat{\mathbf f}$,
$\;\mathbf p \leftarrow \mathbf p + (Z'-Z)\,\hat{\mathbf f}$. Combining the silhouette-derived angular size $\alpha$ with the corrected depth
$Z'$, we finalize the metric scale as: $s \;=\; \frac{\alpha\,Z'}{w_{\text{obj}}} \;=\; \frac{\hat w\,Z'}{f_x\,w_{\text{obj}}}.
$

\topic{Collision Resolution.}
For a colliding pair $(i,j)$, let
$\boldsymbol{\delta}_i,\boldsymbol{\delta}_j$ be the minimal collision-free displacements that separate the pair by moving object $i$ or object $j$, respectively. The two differ because each object is constrained by a different local neighborhood, so its feasible escape direction and distance are object-specific. We apply the smaller least-disruptive move to the corresponding object:
$\mathbf{p}_k \leftarrow \mathbf{p}_k + \boldsymbol{\delta}_k,\quad
k=\arg\min_{k\in\{i,j\}}\lVert\boldsymbol{\delta}_k\rVert$.
If no single move separates them, neighbors in the colliding group are moved as well. When a collision persists, we invoke the VLM to revise the placement order, addressing the conflict at its source rather than locally.

\topic{Reflective Feedback Loop.}
After each placement stage is planned and executed, we render the updated 3D scene and send it back to the VLM together with the input image. The VLM performs a reflective visual check for issues such as incorrect scale, inaccurate orientation, or misplaced items, and applies corrective actions before the pipeline proceeds to the next stage, preventing error accumulation. It also reasons over groups of visually matched objects to equalize their scale. For instance, chairs placed as a matched pair are inferred to share a common size. After the decoration stage, the VLM further counts the placed objects against the
reference and fills any missing instance, either by reusing an existing mesh of the same type or by regenerating one through mesh generation.

\subsection{Lighting}
\label{subsec:relit}
For photorealistic rendering, the VLM assigns materials and recovers the scene's lighting. Since the image-to-3D generator outputs only a baked base color, the VLM infers each object's dominant PBR material from per-category priors, such as a glass table being transmissive. It then identifies the emissive sources such as lamps and windows and estimates each one's activation state, color, and intensity from the reference. Finally, a reflective loop compares the Blender~Cycles render against the reference and refines these parameters until the appearance matches.

%% file: sections/04_experiments.tex
\vspace{-5pt}
\section{Experiments}

\subsection{Experiment Protocol}

\topic{Benchmarks.}
We evaluate our method on HARMONY300, a benchmark we construct using 300 single-image indoor scenes spanning diverse layouts, styles, and lighting conditions. HARMONY300 includes 100 synthetic scenes from Front3D~\citep{front3d,fu20213d}, which provide 3D ground truth, along with 200 copyright-free in-the-wild images. We report perceptual metrics on the full set of 300 scenes and geometric metrics on the 100 Front3D scenes.

\topic{Evaluation Metrics.}
To evaluate the rendering quality of our method, following VIGA~\citep{viga}, we render each method from the input viewpoint and compare against the reference image using image-similarity metrics, including Negative-CLIP score (N-CLIP), which is $1-\text{CLIP}_{\cos}(I_{\text{pred}},I_{\text{ref}})$, where $\text{CLIP}_{\cos}$ denotes CLIP-ViT-B/32 image embeddings and photometric loss (PL), which is the
pixel-wise MSE over normalized RGB in $[0, 1]$. For datasets with ground-truth meshes, \ie, Front3D, we also evaluate the geometry quality and report Chamfer Distance (CD) and F-score at thresholds $0.1$, $0.01$ and $0.001$. 
We additionally conducted a user study with 16 participants across 20 scenes, where participants ranked our method against five baselines. We report the mean rank with standard deviation, as well as the percentage of scenes ranked first (Top-1) or among the top two (Top-2). 

\topic{Baselines.}
We compare \name against representative single image to 3D scene methods:
Gen3DSR~\citep{dogaru2025gen3dsr}, 3D-ReGen~\citep{3dregen},
CAST~\citep{yao2025cast}, SAM3D~\citep{sam3dteam2025sam3d3dfy}, VIGA~\citep{viga}, and the GPT-6 Astra. All methods use their officially released checkpoints where available. Note that SAM3D requires per-object masks as input, so we use \name's segmentation mask and denote the baseline as SAM3D*. 
CAST has no official release, so we use the best available unofficial implementation\footnote{https://github.com/FishWoWater/CAST.}. Also, CAST and SAM3D don't reconstruct backgrounds; we augment with our method's background when calculating perceptual metrics.

\vspace{-7pt}
\begin{figure*}[htbp]
    \centering
    \includegraphics[width=\textwidth]{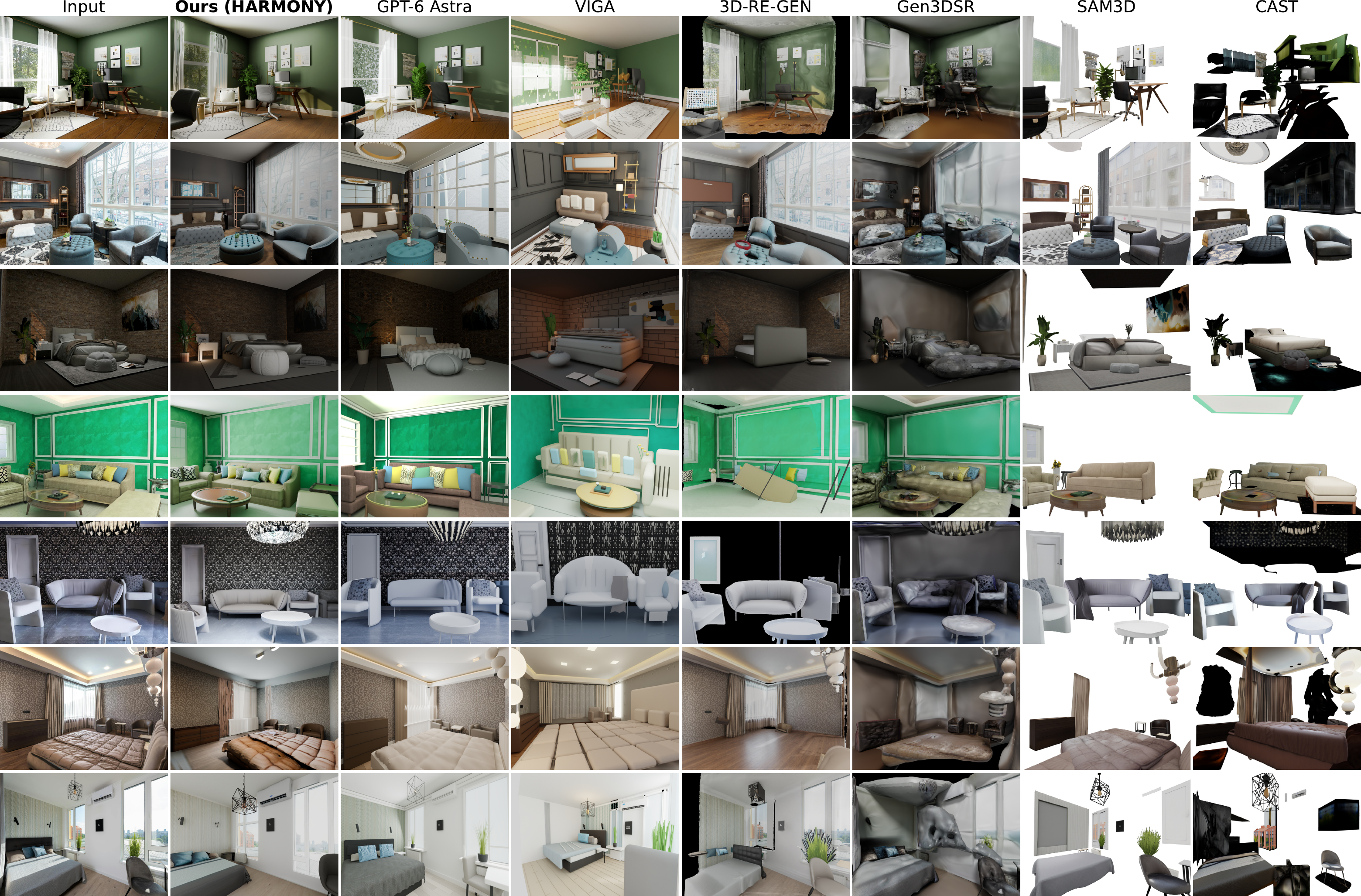}
    \vspace{-15pt}
    \caption{Qualitative comparison between \name and baselines. 
    }
    \label{fig:qualitative}
    \vspace{-12pt}
\end{figure*}

\subsection{Quantitative Results}

\topic{Geometric Evaluation on Front3D.}
\Cref{tab:front3d} reports geometry quality on the 100 Front3D scenes of HARMONY300. \name achieves the lowest Chamfer Distance and the highest F-score at every threshold, since each placement is grounded in dense metric image evidence (silhouettes and depth) and each object is reconstructed with a strong image-to-3D generator. GPT-6 Astra produces visually plausible scenes but exhibits larger geometric errors; for example, beds are often misaligned with walls.


\topic{Perceptual Evaluation on HARMONY300.}
\Cref{tab:userstudy} (left) reports perceptual metrics on all 300 scenes of HARMONY300. \name achieves the best N-CLIP and PL, indicating that its reconstructions most faithfully preserve both the semantic content and the visual appearance of the input image across diverse layouts, styles, and lighting conditions. Notably, baselines that perform competitively on one metric fall behind on the other: Gen3DSR attains a PL close to ours but the worst N-CLIP, suggesting that matching low-level appearance alone does not ensure that the scene content is correctly recovered.

\topic{User Study.}
\Cref{tab:userstudy} (right) summarizes the user study. \name is ranked first in \textbf{67.0\%} of responses and within the top two in \textbf{86.0\%}, with a mean rank of \textbf{1.54}; the next-best method, SAM3D*, reaches only 13.7\% Top-1 with a mean rank of 3.26. This large margin indicates that human judges consistently find our reconstructions more visually and semantically faithful to the input than those of competing methods.

\vspace{-10pt}
\noindent
\begin{minipage}[t]{0.48\textwidth}
\captionof{table}{Geometric results on the 100 Front3D scenes of HARMONY300.}
\vspace{-6pt}
\resizebox{\columnwidth}{!}{
\begin{tabular}{lcccc}
    \toprule
     Method & CD$\downarrow$ & F@0.1$\uparrow$ & F@0.01$\uparrow$ & F@0.001$\uparrow$ \\
    \cmidrule(r){1-1}\cmidrule{2-5}
     SAM3D*      & 0.056 & 81.79 & 14.44 & 0.089 \\
     Gen3DSR     & 0.057 & 78.33 & 10.01 & 0.050 \\
     CAST        & 0.052 & 85.70 & 14.39 & 0.093 \\
     3D-ReGen    & 0.060 & 78.87 & 12.65 & 0.088 \\
     VIGA        & 0.055 & 83.90 & 12.67 & 0.074 \\
     \textbf{Ours} & \textbf{0.049} & \textbf{89.82} & \textbf{19.47} & \textbf{0.130} \\
     GPT-6 Astra & 0.061 & 77.44 & 14.04 & 0.090 \\
    \bottomrule
\end{tabular}
}
\label{tab:front3d}
\end{minipage}%
\hfill
\begin{minipage}[t]{0.48\textwidth}
\captionof{table}{Perceptual results on HARMONY300 (300 scenes) and user study on 20 selected scenes.}
\vspace{-6pt}
\resizebox{\columnwidth}{!}{
\begin{tabular}{@{}lccccc@{}}
    \toprule
     & \multicolumn{2}{c}{Perceptual} & \multicolumn{3}{c}{User Study} \\
    \cmidrule(lr){2-3}\cmidrule(l){4-6}
     Method & N-CLIP$\downarrow$ & PL$\downarrow$ & Mean Rank$\downarrow$ & Top-1 (\%)$\uparrow$ & Top-2 (\%)$\uparrow$ \\
    \midrule
     SAM3D*      & 0.139 & 0.065 & 3.26 $\pm$ 1.60 & 13.7 & 39.3 \\
     Gen3DSR     & 0.218 & 0.054 & 3.34 $\pm$ 1.59 & 11.3 & 37.3 \\
     CAST        & 0.183 & 0.087 & 4.15 $\pm$ 1.44 & 3.3  & 14.7 \\
     3D-ReGen    & 0.146 & 0.089 & 4.34 $\pm$ 1.49 & 4.0  & 13.7 \\
     VIGA        & 0.184 & 0.074 & 4.37 $\pm$ 1.23 & 0.7  & 9.0 \\
     \textbf{Ours} & \textbf{0.107} & \textbf{0.053} & \textbf{1.54 $\pm$ 0.93} & \textbf{67.0} & \textbf{86.0} \\
     GPT-6 Astra & 0.132 & 0.057 & -- & -- & -- \\
    \bottomrule
\end{tabular}
}
\label{tab:userstudy}
\end{minipage}

\subsection{Qualitative Results}
Qualitative comparisons are shown in \Cref{fig:qualitative}. Gen3DSR~\citep{dogaru2025gen3dsr} optimizes objects in the scene context using Score Distillation Sampling (SDS). While it roughly preserves the room layout, its results exhibit substantial geometry and appearance degradation, with nearby objects often merged, support relationships distorted, and textures blurred or flattened. In contrast, our method reconstructs objects individually with generative models, producing higher-quality geometry and textures.
3D-RE-GEN~\citep{3dregen} recovers major furniture and approximate scene layouts, but often produces inaccurate object orientations, scales, and placements due to its reliance on noisy geometric cues. Small objects and decorations are also frequently missing or misplaced, while limited modeling of object relationships can lead to floating or incorrectly supported objects.
VIGA~\citep{viga} relies on VLM-based critique of the final rendered scene to iteratively refine the reconstruction, providing limited direct geometric supervision for individual object placements. In contrast, HARMONY combines a globally grounded floorplan with dense local geometric cues, including silhouettes and depth, to refine each object against explicit geometric targets. SAM3D~\citep{sam3dteam2025sam3d3dfy} relies on accurate per-object segmentation and degrades substantially when applied directly to the full image, limiting its robustness in cluttered scenes. CAST~\citep{yao2025cast} similarly struggles with object segmentation and pose estimation, leading to missing objects and inaccurate spatial configurations.
Overall, HARMONY leverages VLM-based semantic and spatial reasoning to establish coherent scene structure and object relationships, followed by local geometry-based refinement for precise placement. This combination yields more faithful object poses and more coherent spatial arrangements.

\vspace{-10pt}
\subsection{Ablation Study}
\label{sec:ablation}
\Cref{tab:ablation} ablates each pipeline component in isolation, including the
separate roles the VLM plays (reasoning, placement order, refinement) and the
depth-first furniture traversal. Removing silhouette refinement and camera
calibration hurts perceptual similarity the most because the accuracy comes from grounding
placements in dense, metric image evidence. Removing the canonicalized
detection and feedback loop, and the VLM reasoning collapses
semantic fidelity (N-CLIP) toward reasoning-free baselines. The
depth-first-traversal and placement-order ablations isolate structured
reasoning: even with correct per-object estimates, unordered placement causes
occlusion and attachment errors. Our full method benefits from the
\emph{interaction} of global geometric grounding and structured,
image-supervised reasoning, not any single component.
\vspace{-7pt}
\begin{table}[htbp]
\centering
\small
\caption{Ablation of our key design choices on 10 scenes.}
\vspace{-7pt}
\setlength{\tabcolsep}{5pt}
\begin{tabular}{lcclcc}
\toprule
Variant & N-CLIP$\downarrow$ & PL$\downarrow$
& Variant & N-CLIP$\downarrow$ & PL$\downarrow$ \\
\cmidrule(r){1-3} \cmidrule(l){4-6}
\name\ (full)
& \textbf{0.1034} & \textbf{0.0462}
& w/o VGGT refinement
& 0.1069 & 0.0509 \\

w/o depth-first traversal
& 0.1045 & 0.0523
& w/o VLM placement reasoning
& 0.1079 & 0.0522 \\

w/o placement refinement
& 0.1051 & 0.0516
& w/o feedback loop
& 0.1184 & 0.0509 \\

w/o placement order
& 0.1065 & 0.0477
& w/o camera calibration
& 0.1381 & 0.0596 \\
\bottomrule
\end{tabular}
\vspace{-10pt}
\label{tab:ablation}
\end{table}



%% file: sections/05_conclusion.tex
\section{Conclusion and Future Work}
In this paper, we present HARMONY, a hierarchical agentic reasoning framework for reconstructing compositional 3D scenes from a monocular indoor image. Starting from an empty 3D room, HARMONY first calibrates the
camera against the reference image, then places objects in a hierarchical order each stage followed by refinement from geometry-grounded models.
We leverage the strengths of VLMs for spatial reasoning and visual
geometry-grounded models for geometry perception.
A reflective feedback loop after each stage prevents error propagation. Experiments on both synthetic and real-world indoor images show that HARMONY produces compositional 3D reconstructions that align closely with the input. Future works can extend our work to multi-view images and also infer object articulations.


\section*{Acknowledgement}
The authors would like to thank Apple Inc. for supporting this project. The authors would also like to thank Qiao Feng and Minseong Kweon for proofreading this manuscript.

%% file: sections/X_suppl.tex
%
%

\providecommand{\method}{HARMONY\xspace}

\clearpage
\setcounter{page}{1}
\appendix

\etoctoccontentsline{part}{Appendix}
\localtableofcontents

\section{Overview}
\label{sec:arch}
\method reconstructs a scene $\mathcal{S}=(\mathcal{G},\{O_i\})$ with room geometry $\mathcal{G}$ (walls, floor, ceiling, and openings) together with a set of placed objects, each $O_i=(\mathcal{M}_i,T_i,\mathbf{p}_i,\Theta_i)$ carrying a mesh $\mathcal{M}_i$, texture $T_i$, pose $\mathbf{p}_i$, and PBR material factors $\Theta_i$ (roughness, metallicity, index of refraction, and, for transmissive objects, transmission) from a single photograph input $I$. Reconstruction proceeds through six sequential phases, from floorplan and camera recovery, wall-mounted object placement, furniture placement, ceiling object placement, decoration placement, to lighting estimation. Unlike a monolithic
reconstruction network, \method rebuilds every object \emph{independently} through a common set of roles, and gates each stage with object-level and geometric verifiers so that errors are caught and repaired locally rather than propagated downstream.

\section{Implementation Details}
\label{sec:impl}


\subsection{Foundation models}
\method composes off-the-shelf foundation models; no component is trained or fine-tuned. \Cref{tab:models} lists every model and its role. All vision--language reasoning (scene analysis, object verification, layout analysis, decoration comparison, lighting estimation) is served by a single open-vocabulary VLM. All amodal completion and object-appearance synthesis is served by a single image-editing diffusion model. Geometry-to-mesh generation
uses an image-to-3D model.

\begin{table*}[t]
\centering
\small
\setlength{\tabcolsep}{4pt}
\begin{tabular}{@{}lll@{}}
\toprule
\textbf{Model} & \textbf{Role in \method} & \textbf{Access} \\
\midrule
gpt-5.5~\citep{openai2026gpt55release}   & scene / layout / lighting VLM & API \\
Gemini-Flash-2.5~\citep{Gemini2026Image} & amodal completion, appearance & API \\
Hunyuan3D-2~\citep{lai2025hunyuan3d25}              & image-to-3D mesh + texture    & local server \\
VGGT~\citep{vggt}            & Manhattan / metric alignment       & local \\
LocateAnything~\citep{wang2025locateanything}   & open-vocab. box proposals          & local \\
SAM2~\citep{ravi2024sam2}             & instance segmentation              & local \\
Blender (Cycles)& physically-based relighting        & local \\
\bottomrule
\end{tabular}
\caption{Foundation models used by \method. The two prompted models
(VLM and image editor) can be served locally or through hosted APIs with no change to the pipeline.}
\label{tab:models}
\end{table*}

\subsection{Camera Calibration Parameters}
\label{subsec:camera-calib-params}
Camera pose solving uses the following fixed values, held constant across all scenes: the row-band fraction $\rho=0.2$ (fraction of image rows sampled for the floor/ceiling bands); the wall-normal alignment threshold $\tau=0.7$ (minimum $|\text{normal}\cdot\text{axis}|$ for a point to count toward a given bounding plane); and the extension percentiles $p_{\rm fc}=98$ (floor/ceiling) and $p_{\rm wall}=97$ (the two wall planes), used to extend each coarse plane fit outward.

Algorithm~\ref{alg:camera-pose-solving} summarizes our implementation of the camera pose calibration procedure described in the paper.

\begin{algorithm}[t]
\caption{Camera Pose Solving: estimate a Manhattan frame from the VGGT point cloud, refine its six bounding planes, then solve in closed form for the single rigid-body similarity transform that pins the deepest floor/ceiling corner to its canonical wall corner.}
\label{alg:camera-pose-solving}
\begin{algorithmic}[1]
\Require $K,R_0,\mathbf t_0$ (VGGT intrinsic/extrinsic), point cloud with per-point normals; room dims $W_{\rm room},D_{\rm room},H_{\rm room}$; row-band fraction $\rho$, normal threshold $\tau$, percentiles $p_{\rm fc},p_{\rm wall}$
\Ensure $R,\mathbf c$ — calibrated camera rotation and center in the canonical room frame
\State ${\mathbf a_w,\mathbf a_v,\mathbf a_d} \gets \textsc{SVDCluster}(\text{normals})$
\Statex \Comment{Manhattan axes via iterative SVD normal clustering}
\State ${\text{planes}_k},{k\in{w,v,d}} \gets \textsc{HistogramFit}(\text{points},{\mathbf a_k})$
\Statex \Comment{coarse per-axis bounding-plane fit (density peak)}
\For{$k \in {w,v,d}$}
\State extend $\text{planes}_k$ outward to the $p$-th percentile of points with $|\text{normal}\cdot\mathbf a_k|>\tau$
\Statex \Comment{floor/ceiling ($k{=}v$): restricted to the bottom/top row-band of fraction $\rho$; walls ($k{\in}{w,d}$): over the full image}
\EndFor

\State $\{\text{corners}_i\}_{i=1}^{8} \gets
\textsc{BoxCorners}(\{\mathbf a_k\}, \{\text{planes}_k\})$
\State $\overline{\bm X}_f \gets
\arg\max_{\bm X \in \mathcal C_f}
\mathbf e_z^\top(R_0\bm X+\mathbf t_0)$
\label{ln:deepest}
\Statex \Comment{$\mathcal C_f$: floor corners in front of the camera and projecting inside the image}
\Statex \Comment{deepest visible floor corner}
\State $\overline{\bm X}_c \gets$ vertical partner of $\overline{\bm X}_f$ \Comment{same corner, ceiling side}
\State Swap $\mathbf a_w$ and $\mathbf a_d$ if
$|\mathbf a_w^\top R_0^\top\mathbf e_z|
> |\mathbf a_d^\top R_0^\top\mathbf e_z|$
\Statex \Comment{assign width/depth by alignment with camera forward}
\State fix signs of $\mathbf a_w,\mathbf a_d$ so $R_{\rm align}=[\pm\mathbf a_w;\ \mathbf a_v;\ \pm\mathbf a_d]$ has $\det=+1$
\State $s \gets H_{\rm room}/\lVert\overline{\bm X}_c-\overline{\bm X}_f\rVert$ \Comment{metric scale from the anchor edge}
\State $\mathbf t \gets \widehat{\bm X}f - sR_{\rm align}\overline{\bm X}_f$ \Comment{pin floor anchor to its canonical corner}
\State $R \gets R_0 R_{\rm align}^\top$, \quad $\mathbf c \gets sR_{\rm align}\mathbf c_0 + \mathbf t$,\ \ $\mathbf c_0 = -R_0^\top\mathbf t_0$
\State \Return $R,\mathbf c$
\end{algorithmic}
\end{algorithm}

\subsection{Image-to-3D settings}
For each isolated, amodally completed object we generate a textured mesh with Hunyuan3D-2. Mesh fidelity is governed by three parameters: the marching-cubes octree resolution, the number of shape-diffusion steps, and the target face count after decimation. We use an octree resolution of $384$, $25$ diffusion steps, and a target of $80{,}000$
faces; these values noticeably reduce fragmented thin structures (foliage, slats) relative to the model's fast-preview defaults ($128$ / $5$ / $40{,}000$). 

\subsection{Hardware and runtime}
All results were produced on NVIDIA L40 GPUs (46\,GB each). Reconstructing one scene takes $20$--$90$ minutes depending on object count and VLM latency, averaging $3.4$ VLM calls per placed object. A full scene from a single photograph to a furnished, relit \texttt{scene\_full.glb} is completed without any per-scene manual intervention; the pipeline is re-entrant, and each stage skips work already
present on disk.

\subsection{Rendering}
\method exports renderable meshes compatible with any renderer; before it enters the relighting stage, \method utilizes \texttt{pyrender} under flat ambient light using each mesh's baked-in texture, while the relighting stage (\cref{sec:lighting}) relights with Blender's Cycles.

\subsection{Baseline Setup}
\label{sec:baselines}
For every baseline we use the officially released code (except for CAST which has no official code), feed the same monocular image, and render the resulting scene from the camera each method recovers. Retrieval- and part-assembly
baselines are rendered with their native exporters; mesh-only baselines are
rendered under the \emph{same} \texttt{pyrender} flat-ambient setup as
\method (\Cref{sec:impl}) so appearance differences reflect reconstruction,
not shading. All baseline outputs in our qualitative comparison (\Cref{sec:qual}) are produced by this uniform protocol.

We run Gen3DSR with the authors' stock
default configuration. VIGA and 3D-RE-GEN
required non-default changes to run on our evaluation set at all or to stay
computationally tractable at our scale; we list them here for
reproducibility. For VIGA, we use SAM3D for object reconstruction and \texttt{GPT-5.5} for the Blender code agent. 
For 3D-RE-GEN, we use \texttt{gpt-image-2} for the background empty-room inpainting instead of their default Gemini-Image, which we found \texttt{gpt-image-2} works much more reliably.
SAM3D requires precise per-object segmentation and degrades on
whole-image input, so we augment it with each object's mask from HARMONY's own segmentation
colormap (\Cref{sec:eval}).

\section{Failure Cases}
\begin{figure*}[htbp]
    \centering
    \includegraphics[width=\textwidth]{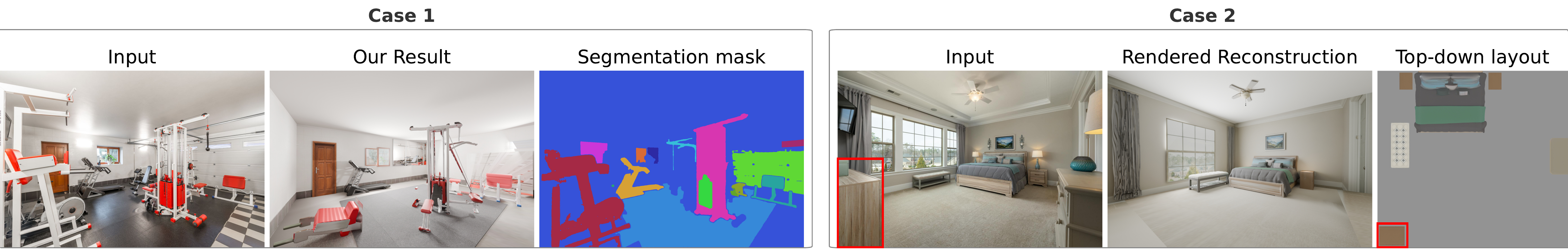}
    \vspace{-15pt}
    \caption{Failure cases of \name. Case 1 (gym) shows rare items that fail to detect or inpaint; In Case 2, the red box marks an object present in the input image and reconstructed scene in top-down layout but occluded in the rendered reconstruction, since it sits against a wall occluded from render camera. 
    }
    \label{fig:failure-cases}
    \vspace{-12pt}
\end{figure*}

We discuss our three common failure cases here and examples can be found in \Cref{fig:failure-cases}.


\topic{Case 1: Rare-type synonym mismatch.} For rare object types, a mismatch between the VLM's and the detector's vocabulary would affect detection: $25.9\%$ of detections carry a low grounding confidence ($<\!0.35$), and genuine placement failures (after routing and de-duplication are excluded) remain rare at $0.3\%$ of objects. These issues can be resolved using stronger foundation models within the same framework.

\topic{Case 2: Heavy occlusion due to foreground clipping.} When an object sits very close to the camera and is largely clipped, the VLM still recovers the correct semantic relation (e.g., which wall it leans against), but the reconstructed mesh is not well constrained from the render viewpoint.

\section{Prompts for Each Stage}
\subsection{Geometry Preprocessing}
\label{sec:asset}
\begin{figure*}[t]
  \centering
  \includegraphics[width=\textwidth]{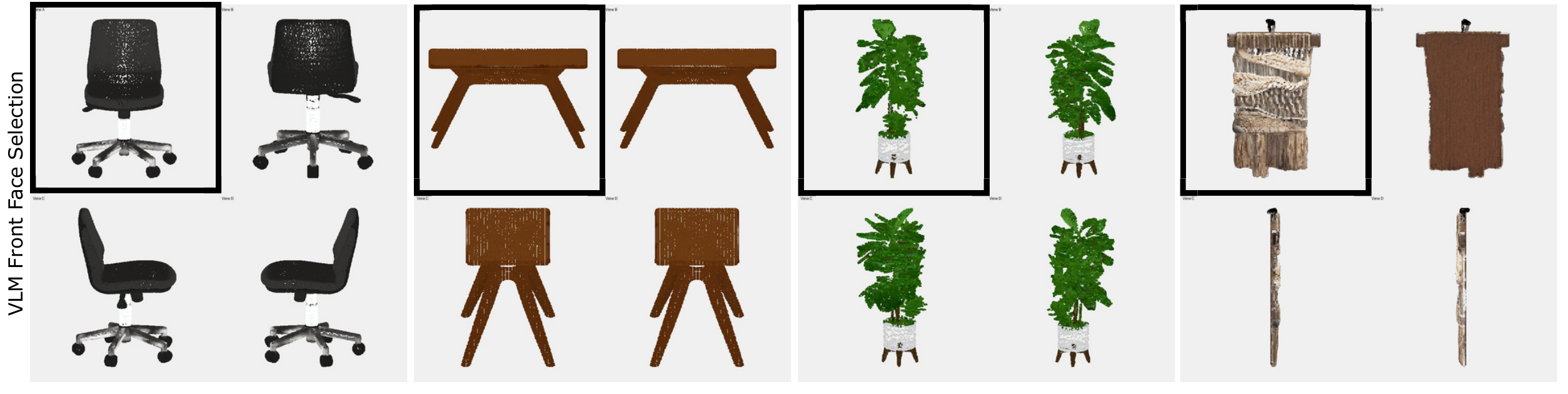}
  \vspace{-15pt}
  \caption{Visualization of VLM-selected front face for each object for
  PCA-aligned meshes.}
  \vspace{-10pt}
  \label{fig:canonicalization}
\end{figure*}

\topic{Canonicalization} The VLM-selected front face for each canonicalized mesh is shown in
\Cref{fig:canonicalization}; the abridged prompt is below:

\begin{tcolorbox}[
    breakable,
    colback=gray!10,    
    colframe=gray!80,   
    fontupper=\ttfamily, 
    boxrule=0.5pt,      
    arc=2mm,            
    left=2mm,           
    right=2mm,          
    top=1mm,            
    bottom=1mm          
]
\par\noindent{\small\ttfamily
Given a 2x2 grid of the de-tilted mesh rendered at four yaw rotations
(A=0deg, B=90deg, C=180deg, D=270deg), pick the panel showing the
object's FRONT, using per-category hints (e.g.\ a sofa/chair faces its
seat concavity).\par\smallskip
OUTPUT -- JSON ONLY: \{"front\_panel": "A|B|C|D"\}.
\par}
\end{tcolorbox}

The chosen panel's yaw is baked in permanently: the mesh is rotated by
that yaw (after PCA de-tilt), re-centered, and re-exported as the
object's canonical GLB, with the resulting \texttt{front\_local\_axis}
saved to \texttt{detilt\_results.json} for the placement stage to
consume.

\begin{figure*}[htbp]
  \centering
  \includegraphics[width=\textwidth]{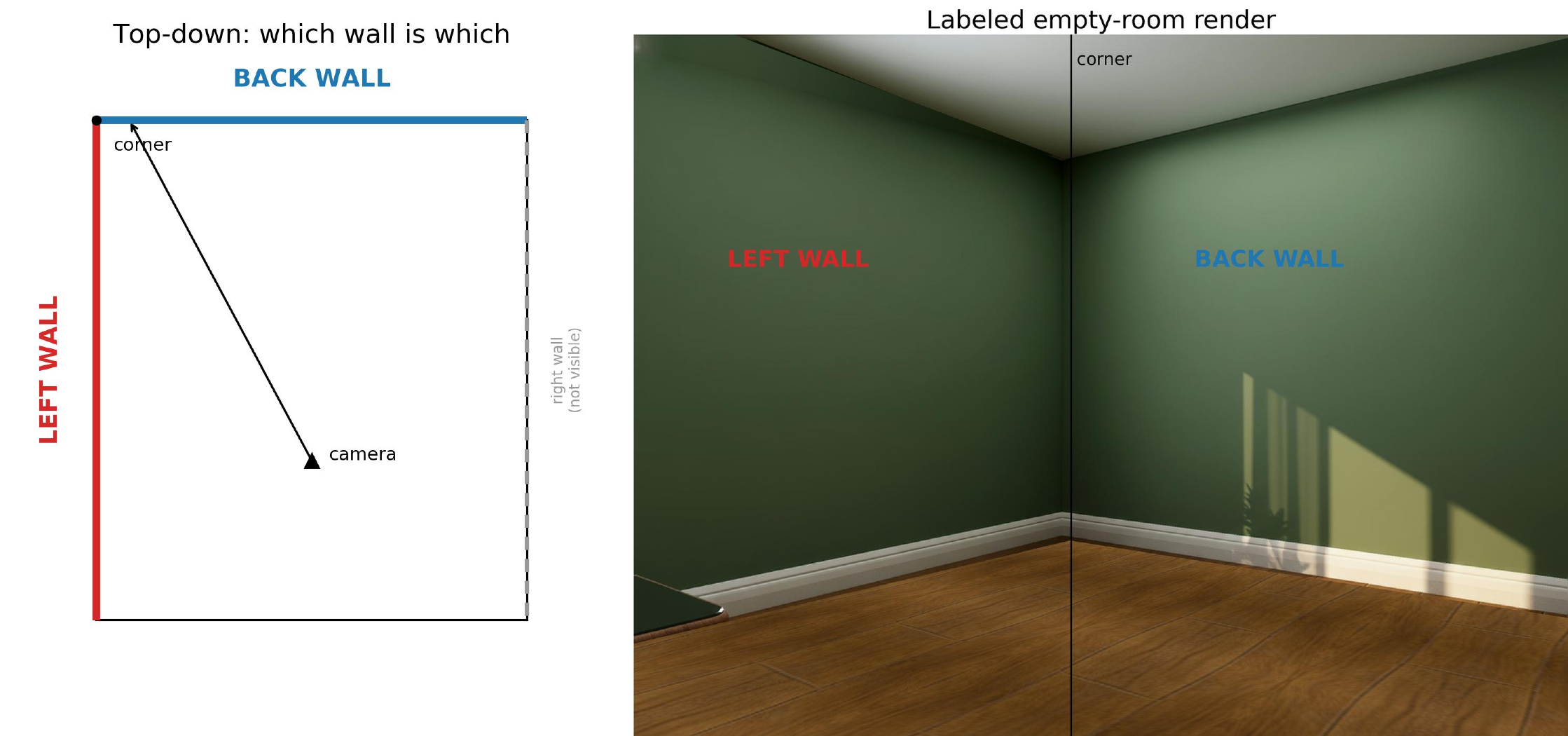}
  \caption{Wall labeling example. Left: top-down
floor plan with the camera position and viewing direction. Right: the corresponding reference-textured empty-room render, with the same
left/back wall labels and their shared corner marked.}
\vspace{-10pt}
  \label{fig:wall-labeling}
\end{figure*}

\topic{Layout Generation} The layout stage recovers room geometry and camera pose from the single input
photograph. The VLM analyzes the image in a fixed step order, tracing the
baseboard, counting openings, inferring room type and dimensions, recovering camera pose and returning a strict JSON description of the room, its walls and
openings, the camera, and per-surface texture descriptions; the camera is
then refined by corner-pinning and analytic orbit correction against a
render of the recovered empty geometry (\Cref{subsec:3d-room-layout}). \Cref{fig:wall-labeling} shows an example: the top-down floor plan on the
left and the corresponding empty-room render on the right, with the same
wall labels marked on both, and the corner where they meet.
Throughout this appendix,
runtime-substituted values are shown in
braces (e.g.\ \texttt{\{room\_width\}}); decorative Unicode rules in the
originals are rendered here as ASCII, and the complete byte-exact templates are
released with the code. The following prompt shows the step skeleton and output
schema.

\begin{figure*}[htbp]
  \centering
  \includegraphics[width=\textwidth]{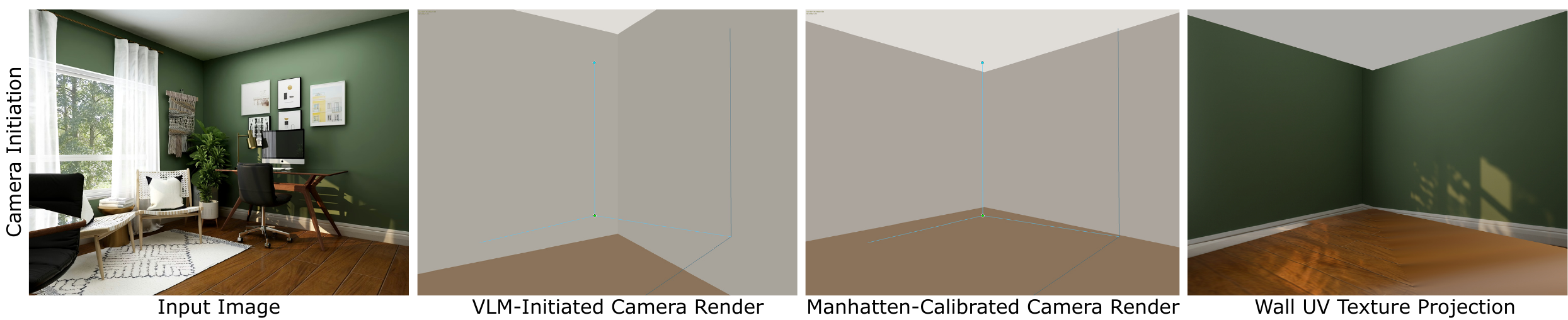}
  \vspace{-10pt}
  \caption{Visualization of the calibration process during layout generation stage.}
  \label{fig:layout_gen}
\end{figure*}

\begin{tcolorbox}[
    breakable,
    colback=gray!10,    
    colframe=gray!80,   
    fontupper=\ttfamily, 
    boxrule=0.5pt,      
    arc=2mm,            
    left=2mm,           
    right=2mm,          
    top=1mm,            
    bottom=1mm          
]
\par\noindent{\small\ttfamily
\par\smallskip
You are reconstructing only what is VISIBLE in a single photograph.
Do NOT infer, guess, or add anything not directly visible.\par\smallskip
STEP 1 -- TRACE THE BASEBOARD. For each continuous baseboard segment,
label the wall (left/back/right) and its run direction; a direction
change is a real room corner (apply the baseboard corner test).\par\smallskip
STEP 2 -- IDENTIFY ROOM TYPE, set dimension priors (floor W x D,
ceiling height) from visible furniture and scale anchors (door
$\sim$0.9m, desk $\sim$0.7m), then estimate each visible wall's
length/height and any opening's width/height/offset (offset+width
$\leq$ wall length).\par\smallskip
STEP 3 -- CAMERA PLACEMENT: height, floor\_junction\_y\_frac (primary
tilt cue), yaw, aim\_target\_world\_m + landmark, dist\_to\_back,
facing, deepest-corner column.\par\smallskip
STEP 4 -- TEXTURES: describe floor/wall MATERIAL only (no
lighting/shadows) and a physical tile\_size\_m for each.\par\smallskip
OUTPUT -- VALID JSON ONLY: room \{room\_type, floor\_width\_m,
floor\_depth\_m, ceiling\_height\_m\}; walls [\{orientation, length\_m,
height\_m, surface\}]; camera \{height\_m, tilt\_deg,
floor\_junction\_y\_frac, yaw\_deg, dist\_to\_back\_m, facing,
corner\_px, back\_wall\_side, aim\_target\_world\_m,
aim\_target\_landmark, visible\_wall\_height\_m,
visible\_floor\_depth\_m\}; floor\_texture \{material, color, pattern,
tile\_size\_m, synthesis\_prompt\}; wall\_texture \{material, color,
finish, tile\_size\_m, synthesis\_prompt\}; blocked\_corners [].
\par}
\end{tcolorbox}

\begin{figure*}[htbp]
  \centering
  \includegraphics[width=\textwidth]{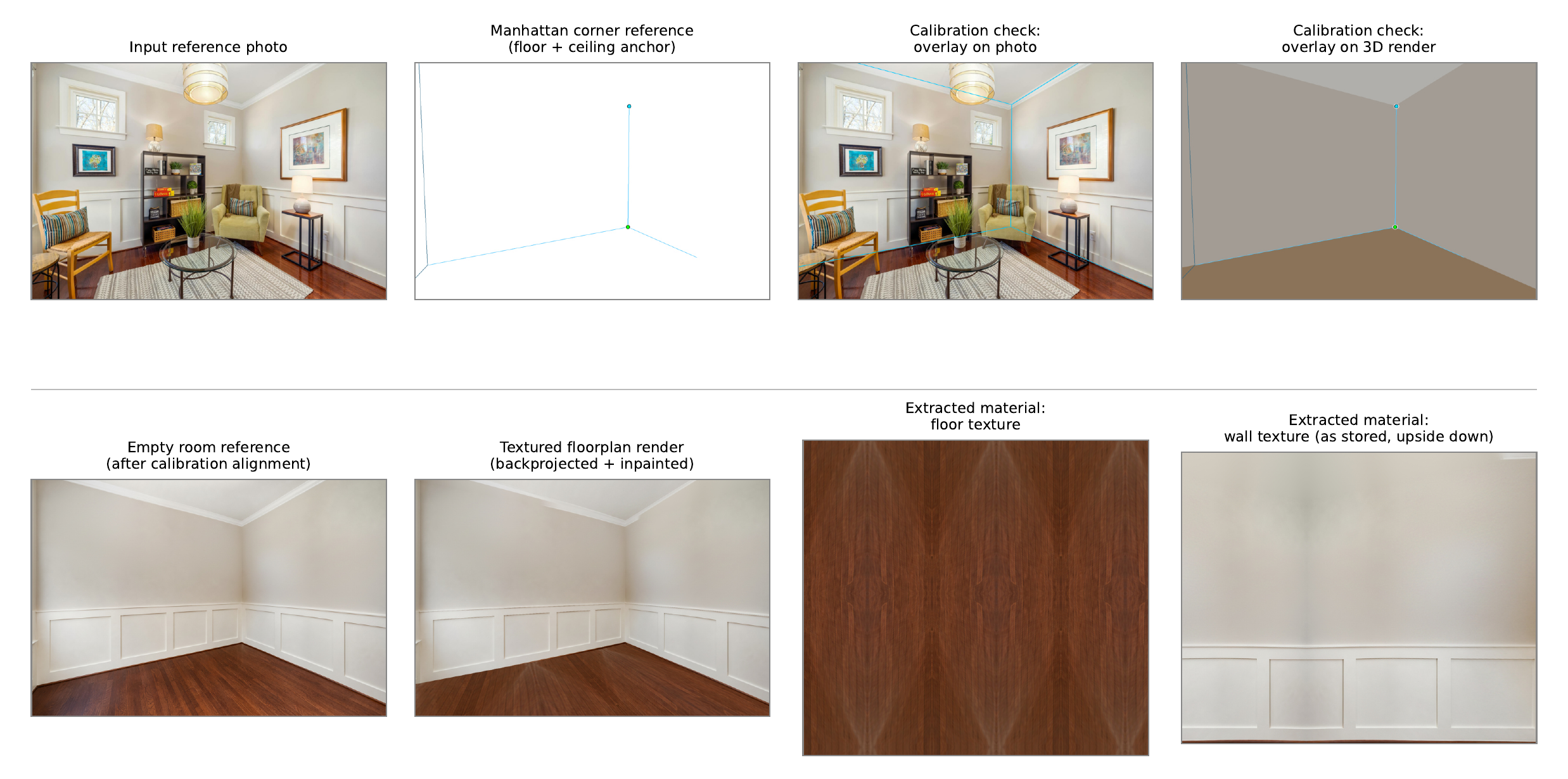}
  \vspace{-20pt}
  \caption{Visualization of the layout initialization stage, left to right and top to bottom. Given a reference image, the Manhattan corner estimated from the converted VGGT point cloud backprojects onto the render plane and shows clear orthogonal structure, with its floor and ceiling anchors and the edges extending from the floor corner. The camera initialized on the canonical wall-box mesh by the VLM is then aligned to this same floor corner, its extending edges, and the ceiling point during camera calibration, anchoring itself onto the backprojected reference until convergence -- shown as the corner reprojected onto both the reference photo and the calibrated 3D render. The empty-room image is inpainted by Gemini-Flash-2.5 from the reference image to remove furniture while preserving room structure, then backprojected onto the aligned floorplan to give it partial texture. Gemini then completes the floor's and each wall's texture into a full, coherent tileable material.}
  \label{fig:layout_gen1}
\end{figure*}

\subsection{Wall-Mounted Object Placement}
\label{sec:wall}
This stage operates on the walls already labeled left/back/right during
layout initialization. Wall-mounted objects, such as windows, doors, curtains, and wall art, are proposed by VLM, detected by LocateAnything and segmented by SAM2, and verified by the VLM, amodally completed with their glass and artwork preserved, rectified to the wall plane for frame-like objects like windows and paintings and doors, meshed, and mounted. Detection emits a structured list; completion runs on the image editor. 

\topic{Rectification and meshing.} For frame-like, planar objects only, such as windows, doors, paintings, and mirrors, the segmented crop is unwarped before meshing: a perspective transform maps its tilted quad (fit with a rotated bounding rectangle) to a fronto-parallel rectangle, so the mesh is generated flat rather than keystoned by the viewing angle. Non-planar objects (shelves, sconces, TVs) skip this step and are meshed directly from the raw crop.

\topic{Placement} 
Position comes from pure geometric back-projection, not a
depth network: the segmentation mask's centroid is cast as a ray from the
camera and intersected against the room's four wall planes (already fixed by
the layout stage) to get both the 3D hit point and which wall it belongs to;
a VLM only adjudicates ties when the ray lands near a corner seam. Real-world
size follows from the mask's pixel extent divided by focal length and scaled
by that hit's camera-space depth, with a foreshortening correction and a
plausibility clamp against per-category default aspect ratios. Orientation
prefers a VLM-identified front face when available, otherwise snaps to the
nearest $90^\circ$ increment of the wall's outward normal, before the mesh is translated to the hit point and snapped flush against the wall.

\topic{VLM Detection Prompt}

\begin{tcolorbox}[
    breakable,
    colback=gray!10,    
    colframe=gray!80,   
    fontupper=\ttfamily, 
    boxrule=0.5pt,      
    arc=2mm,            
    left=2mm,           
    right=2mm,          
    top=1mm,            
    bottom=1mm          
]
\par\noindent{\small\ttfamily
\par\smallskip
List every wall-mounted/wall-attached object clearly visible (window,
door, curtain, shelf, tv, painting, mirror, clock, sconce, radiator,
air conditioner, vent, etc.). A sconce's bracket must be bolted to the
wall itself (excludes floor/table lamps); a curtain hangs from a
wall-mounted rod. Objects resting ON furniture (a TV on a console,
items on a shelf) belong to the decoration stage, not here -- list
only things fixed to the bare wall.\par\smallskip
OUTPUT -- JSON array of strings, e.g.\ ["window", "curtain",
"art"].
\par}
\end{tcolorbox}

\begin{figure}[htbp]\centering
\includegraphics[width=\linewidth]{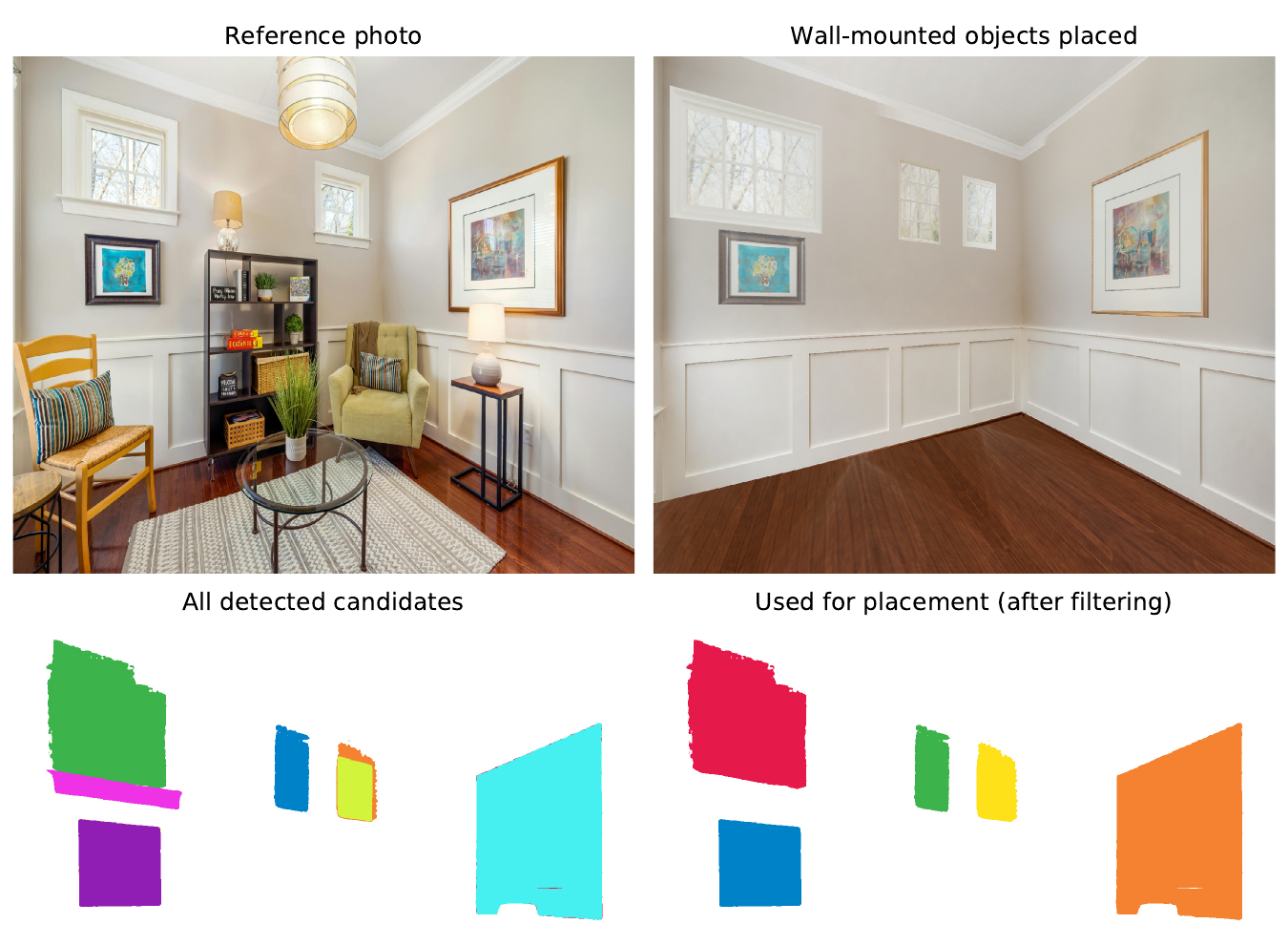}
\caption{Wall-mounted detection-to-placement pipeline. Top: the reference photo and the final render with wall-mounted objects placed. Bottom: all detected segmentation
candidates before filtering, and the subset actually used for placement after containment/fragment removal and verification.}
\label{fig:wallmounted-pipeline}
\end{figure}

\topic{Containment and fragment removal} 
Raw detections are filtered in
two passes before verification. A pre-segmentation box filter drops a
detected box when another box of equal or higher confidence covers most
of its area. The same mechanism also resolves a "gallery wall" box that contains several individual frames by keeping the individuals and dropping the group box. After segmentation, same-type masks with high IoU or near-full containment are collapsed, discarding the lower-confidence duplicate. A separate VLM verification pass then rejects fragments and false positives, such as reflections, a painting-within-a-painting, a ceiling light mistaken for a wall sconce, a sliver at the frame edge, and a rejected candidate is dropped before its mask is ever written to disk, rather than kept with a
failing tag. \Cref{fig:wallmounted-pipeline} shows this end to end for one
scene: all detected candidates, the subset that survives filtering and
verification, and the resulting placement.

\subsection{Furniture Placement}
\label{sec:furniture}
We specify the usage of VLM reasoning here; the usage of silhouette and depth refinement is specified in \cref{subsec:recon}.

\topic{Detection Prompt}


\begin{tcolorbox}[
    breakable,
    colback=gray!10,    
    colframe=gray!80,   
    fontupper=\ttfamily, 
    boxrule=0.5pt,      
    arc=2mm,            
    left=2mm,           
    right=2mm,          
    top=1mm,            
    bottom=1mm          
]
\par\noindent{\small\ttfamily
\par\smallskip
List only floor-standing furniture and floor coverings clearly visible (sofa, chair, table, bed, cabinet, carpet, etc.) -- exclude
anything small or decorative (pillows, books, plants on a shelf). For
each, give a short grounding phrase for open-vocabulary
detection.\par\smallskip
OUTPUT -- JSON array: [\{type, gdino\}], e.g.\ [\{"type":"sofa",
"gdino":"leather sofa . leather couch"\}].
\par}
\end{tcolorbox}

\topic{Depth and Relational Analysis Prompts}
In order to avoid reasoning about dense collisions between multiple placed
objects, depth analysis is applied to ease it at the source. Wall affinity is
assigned relative to the same left/back/right wall labels established during
layout initialization (\Cref{fig:wall-labeling}).

Given the annotated empty-room render, the original photo,
and per-object crops, the VLM then assigns each object a relational description, such as
wall affinity, facing, functional group, and support relations, together
with a dependency-aware placement order and a matched-set count. The placement
solver then snaps each mesh to its reference silhouette and refines it against
a predicted depth map. The box below abridges the layout template.

\begin{tcolorbox}[
    breakable,
    colback=gray!10,    
    colframe=gray!80,   
    fontupper=\ttfamily, 
    boxrule=0.5pt,      
    arc=2mm,            
    left=2mm,           
    right=2mm,          
    top=1mm,            
    bottom=1mm          
]
\par\noindent{\small\ttfamily
\par\smallskip
Room W x D x H and camera pose are given; camera faces the back wall
(Z=0). For every detected object decide: verify (exclude outdoor
scenery / wall-mounted items / on-surface decorations / fragments of a
larger piece; correct wrong type labels); wall\_affinity
(back/left/right/centre -- the object's BACK touches that wall);
relations (on\_top\_of / in\_front\_of / behind / facing\_toward, by
index -- "in front of" means between the anchor and the camera); and
count (total identical instances forming one matched set).\par\smallskip
PLACEMENT ORDER: deepest corner first; wall-affine objects before centre
objects; farther before nearer; dependencies always after their
anchor; include ALL.\par\smallskip
OUTPUT: \{"placement\_order": [\{index, type, exclude, depth,
wall\_affinity, wall, on\_top\_of, in\_front\_of, facing\_toward,
behind, opening\_relation, group, notes, count\}], "scene\_summary":
"..."\}
\par}
\end{tcolorbox}

\topic{Post-placement group refinement}
The post-placement step corrects three group-level errors. First, per-chair orientation is set independently by the analysis mechanisms above and never checked for group coherence, so a dining set can end up with one chair facing the table while its neighbours face slightly outward instead of all pointing inward toward a shared centre. The post-refinement step modifies the orientations of chair groups with anchored table coordinate with visually-inspected correspondence.
Second, the representative-mesh swap copies geometry but not scale, so a group assembled from differently-sized source crops can render with
uniform detail but inconsistent size across members, which is unified by the post-processing step. Third, the post-refinement step analyzes
colour/appearance: if two visually distinct chairs (e.g.\ one light, one dark) are placed on the wrong sides of a symmetric arrangement, it swaps them and re-renders the group.

\subsection{Ceiling Object Placement}
\label{sec:ceiling}
Ceiling-mounted fixtures such as pendants, chandeliers, flush-mount and recessed lights, track lights, and fans, are detected by the VLM, amodally completed (viewed from below), meshed, and hung from the ceiling plane.

Detection and placement both follow the same overall pattern as furniture (\Cref{sec:furniture}) and wall-mounted objects (\Cref{sec:wall}); only the differences are described below.

\topic{Placement} Position and size use the same back-projection-ray and pixel-extent/focal-length/depth estimate as furniture and wall-mounted placement, here intersected with a single
ceiling-height plane rather than a wall or floor; a ray that never reaches this plane instead initializes the object at the room's center and gradually shifts it toward its estimated depth position.

\topic{Decoration Placement}
\label{sec:deco}
Small items are added by comparing, per furniture piece, the reference crop against the clean reconstructed piece, and listing what rests on the surface in the photo but is absent from the reconstruction. Presence and already-placed
checks guard against hallucinated or duplicated decorations before any mesh is generated.

\topic{Missing-decoration detection} 
For each already-placed furniture piece, the VLM compares a reference crop (the piece boxed in red, neighbouring furniture boxed in blue to exclude their items) against the isolated,
furniture-only reconstruction, and lists every object resting on its surface in the photo that is absent from the reconstruction.

\begin{tcolorbox}[
    breakable,
    colback=gray!10,    
    colframe=gray!80,   
    fontupper=\ttfamily, 
    boxrule=0.5pt,      
    arc=2mm,            
    left=2mm,           
    right=2mm,          
    top=1mm,            
    bottom=1mm          
]
\par\noindent{\small\ttfamily
\par\smallskip
IMAGE 1 is a zoomed reference crop with the target furniture boxed in RED and neighbouring furniture boxed in BLUE (objects on blue-boxed
furniture are ignored); IMAGE 2 is the isolated, furniture-only 3D reconstruction of the same piece. List every object resting on the RED piece's surface in IMAGE 1 that is absent from IMAGE 2 (e.g. the surface begins at the top edge of the red box, so items such as plants, lamps, or vases that extend above it count8).\par\smallskip RULES: include any surface item (pillows, blankets, books, plants, lamps, electronics, tableware, etc.) except structural parts (legs, armrests); describe each by quantity, colour, material, and position; trust what is visually seen over the furniture-type
label; use the room type to disambiguate ambiguous small objects; never hallucinate an object that is not visible; return an empty list
if nothing is missing.\par\smallskip
OUTPUT: \{"label": "...", "missing\_decorations": [\{"name", "description"\},
...]\}
\par}
\end{tcolorbox}

\topic{Placement} 
A separate VLM call first matches the decoration to its supporting furniture instance by type, and by size/camera-distance cues when several same-type pieces exist (e.g.\ the larger, closer coffee table vs.\ the smaller, farther one); a post-placement verification pass re-checks
this match against the reference and reassigns it if it landed on the wrong piece, but not for minor position differences within the same piece.
Another VLM call estimates the decoration's real-world size from common-sense knowledge of its category, a coarse placement type (resting flat on the surface, leaning against a backrest like a sofa pillow, or standing on the floor beside the furniture), a left/center/right surface position, a front/middle/back depth position (for on-surface items), and whether it has a meaningful front face that must face into the room (a monitor, a picture frame) or none (a lamp, a pillow).

\begin{tcolorbox}[
    breakable,
    colback=gray!10,    
    colframe=gray!80,   
    fontupper=\ttfamily, 
    boxrule=0.5pt,      
    arc=2mm,            
    left=2mm,           
    right=2mm,          
    top=1mm,            
    bottom=1mm          
]
\par\noindent{\small\ttfamily
\par\smallskip
Given the decoration crop and its supporting furniture: estimate its
real-world width x height x depth in metres from common knowledge (e.g.\ a
standard monitor is roughly 0.55 x 0.45 x 0.22 m); classify placement type
(on\_surface / against\_back / on\_floor); surface position (left/center/right) and, for on\_surface items, depth position
(front/middle/back); and whether it has a meaningful front face that must face toward the room (a screen, a picture frame) or none (a lamp, vase, pillow).\par\smallskip
OUTPUT: \{"size\_m": \{width, height, depth\}, "placement\_type", "surface\_position", "depth\_position", "facing", "reasoning"\}.
\par}
\end{tcolorbox}

\topic{Orientation correction.} The placed object is rendered at
its current pose and at the three other axis-aligned yaw rotations
($90^\circ$ CCW, $180^\circ$, $90^\circ$ CW), the four renders are stitched
into a 2$\times$2 grid, and the VLM picks the one panel whose orientation
(the lamp's arm/head, a screen's face, a book's spine) matches the
reference in a single call. Symmetric
objects (a vase, a centred lamp shade) default to the unrotated panel. Up to
three further single-step fine-tune passes then correct any residual error
the picked yaw didn't fully resolve. 

\topic{Matched-group reordering} For a set of same-type decorations placed together (e.g. pillows across a sofa), the VLM compares the placed left-to-right sequence against the reference's left-to-right sequence by colour and shape, and returns a list of index swaps to correct any ordering mismatch or copy an existing similar mesh to correct the missing object,
re-rendering after each is applied.

\begin{figure*}[t]
  \centering
  \includegraphics[width=\textwidth]{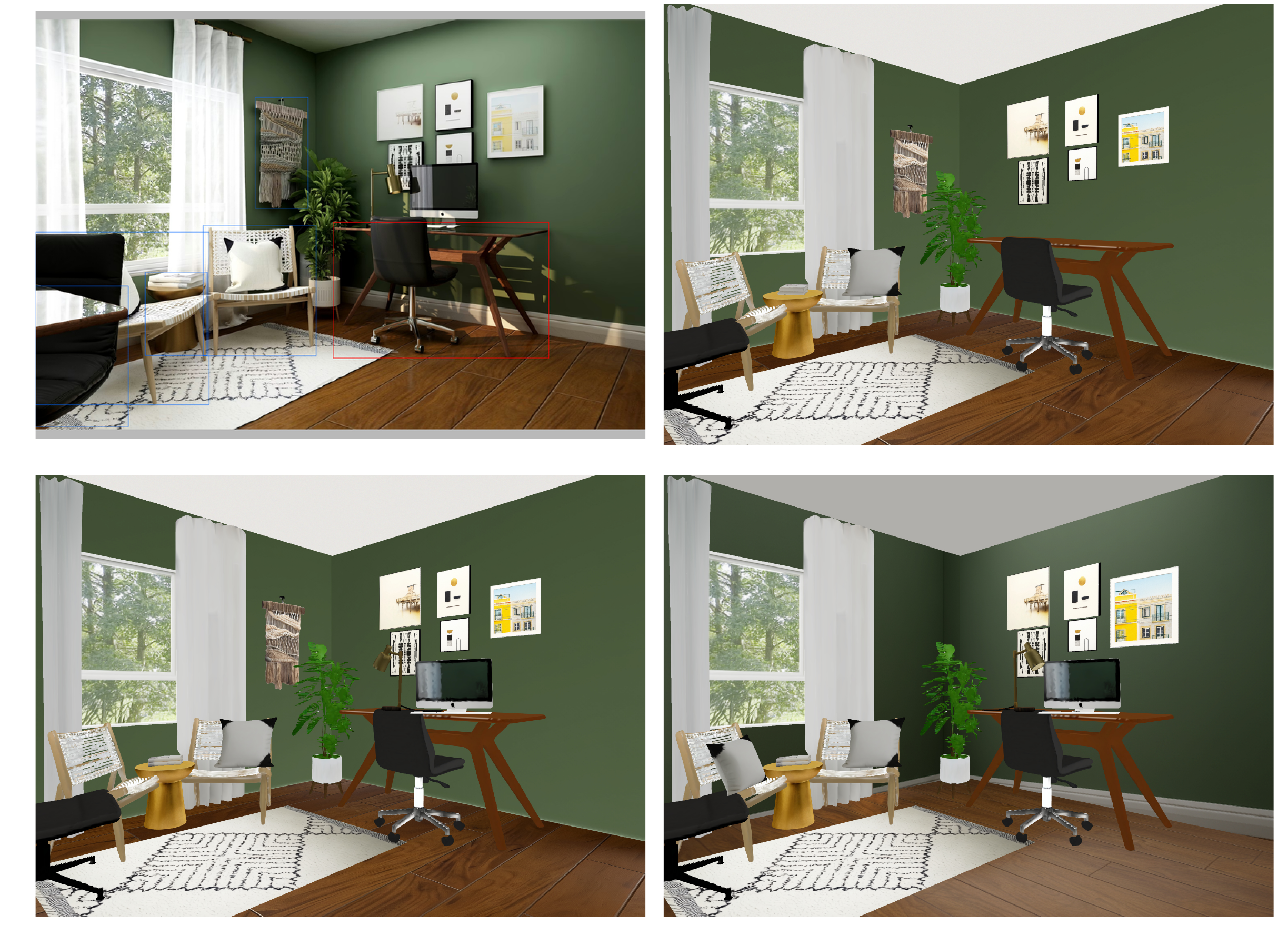}
\caption{\textbf{Decoration Placement Correction.} Top left: Reference image with VLM-selected furniture to reason and place decorations; Top right: Base render before placing decorations onto the desk for the VLM to compare and reason the decorations that should be placed; Bottom left: Initial placement of decorations onto corresponding supporting furniture; Bottom right: \texttt{pyrender} rendering of the scene after per-object orientation correction of the lamp and matched-group reordering of the pillows.}
  \label{fig:deco-fix}
\end{figure*}


\subsection{Material and Lighting Estimation}
\label{sec:lighting}

\topic{Material estimation} For every generated object (furniture,
wall-mounted, ceiling, and decoration GLBs alike), the VLM is shown the
isolated object crop and classifies its dominant surface material into one of
a fixed set of categories, then estimates PBR factors (roughness, metallic,
IOR, transmission) written onto the GLB material and physical properties
(weight, thickness, elasticity) written to a sidecar for downstream physics;
per-category defaults fill in and clamp anything the VLM omits or returns out
of range.

\begin{tcolorbox}[
    breakable,
    colback=gray!10,    
    colframe=gray!80,   
    fontupper=\ttfamily, 
    boxrule=0.5pt,      
    arc=2mm,            
    left=2mm,           
    right=2mm,          
    top=1mm,            
    bottom=1mm          
]
\par\noindent{\small\ttfamily
\par\smallskip
The image shows a single object (\{obj\_type\}) isolated on a plain grey
background, with its real-world bounding size. Estimate its DOMINANT surface
material and physical properties: material category (fabric, leather, wood,
metal, glass, plastic, ceramic, stone, rattan, foam, paper, other),
roughness, metallic, IOR, transmission (0 opaque -- 1 clear glass/acrylic),
weight\_kg, thickness\_m, and elasticity (0 rigid -- 1 springy). Judge from
visible sheen/reflections; if the object mixes materials, report the one
covering the most surface area.\par\smallskip
OUTPUT: \{"material\_category", "roughness", "metallic", "ior",
"transmission", "weight\_kg", "thickness\_m", "elasticity"\}
\par}
\end{tcolorbox}

\topic{Lighting estimation} Given the reference photo and the
world-space positions of every placed light source, the VLM decides which
sources are emitting and their color and intensity, returning a compact
lighting specification consumed by the renderer (shown below).

\begin{tcolorbox}[
    breakable,
    colback=gray!10,    
    colframe=gray!80,   
    fontupper=\ttfamily, 
    boxrule=0.5pt,      
    arc=2mm,            
    left=2mm,           
    right=2mm,          
    top=1mm,            
    bottom=1mm          
]
\par\noindent{\small\ttfamily
\par\smallskip
IMAGE 1 is the reference photo; IMAGE 2 is the current lit preview under
the CURRENT LIGHTING SETUP given below. Diagnose
brightness (too\_bright / matches / too\_dim) and tone (too\_warm /
matches / too\_cool) versus the reference. If too\_dim, raise
\texttt{global\_intensity\_mult} (bounded [0.5, 1.3]) to scale every active
light's intensity, including ambient; if too\_bright, lower it. Optionally
fine-tune individual lights (intensity multiplier, colour shift, on/off)
when only one lamp/window is off; if it already matches, return empty
deltas and verdict "converged".\par\smallskip
OUTPUT: \{"brightness\_assessment", "tone\_assessment",
"global\_intensity\_mult", "ambient\_delta": \{"color\_delta",
"intensity\_mult"\}, "directional\_delta": [\{"index", "color\_delta",
"intensity\_mult"\}], "point\_light\_deltas": [\{"id", "set\_on",
"color\_delta", "intensity\_mult", "offset\_delta\_m"\}],
"window\_deltas": [\{"id", "set\_daylight\_on", "color\_delta",
"intensity\_mult"\}], "verdict"\}
\par}
\end{tcolorbox}

\section{HARMONY300 Dataset}
\label{sec:eval}
\begin{figure*}[t]
  \centering
  \includegraphics[width=\textwidth]{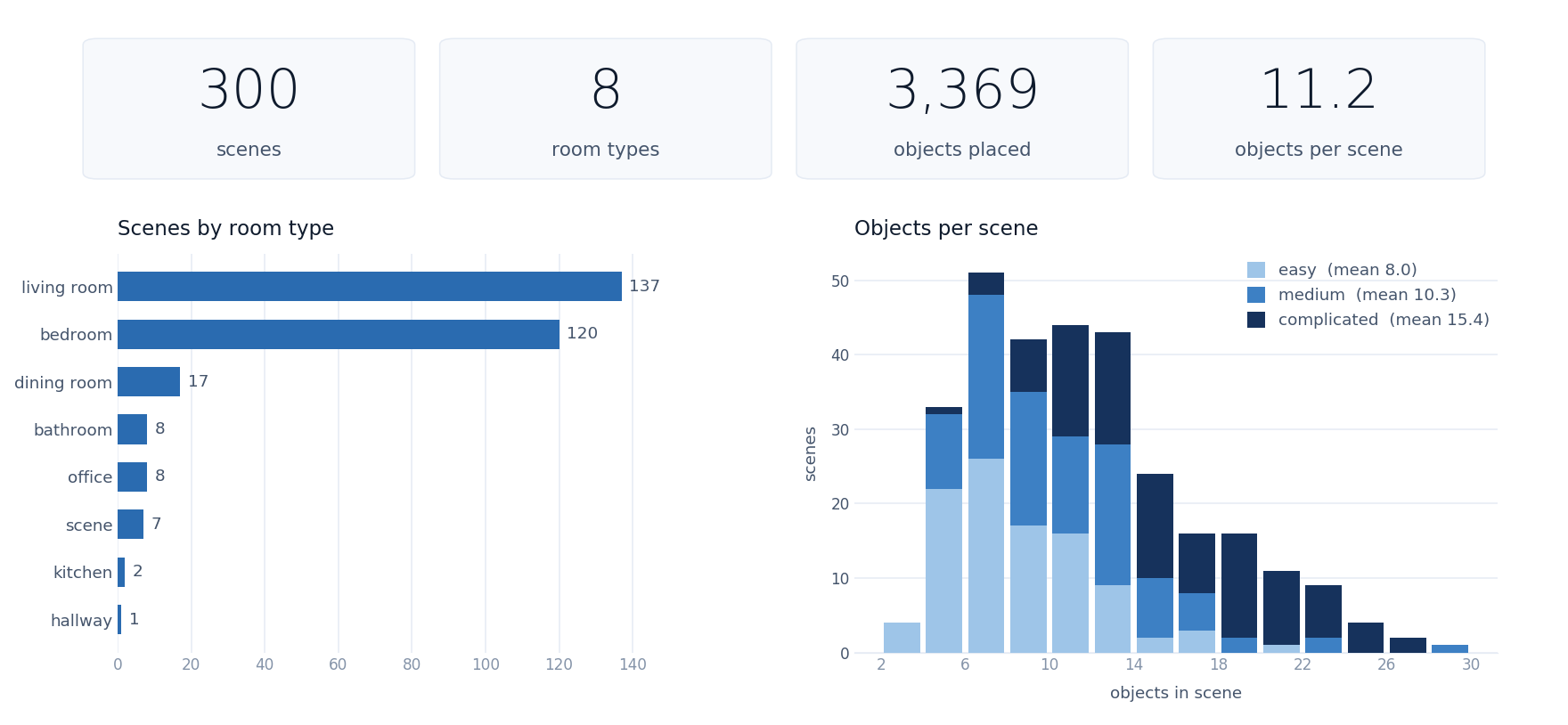}
\caption{HARMONY300 example: reference
photo, instance segmentation colormap, and \method's physically-lit render.}
\label{fig:harmony300_stats}
\end{figure*}

Beyond the curated real-image set presented in the main paper, we release
\textbf{HARMONY300}, a benchmark comprising $300$ single-image indoor scenes
across three difficulty levels. Each scene is paired with its reconstructed
3D scene and a physically lit render produced by \method. \Cref{fig:harmony300_stats}
summarizes the distribution of room types and the number of objects in the
scenes across the dataset.

\subsection{Splits} 
\texttt{easy} ($100$ scenes) are 3D-FRONT synthetic
renders with 3D ground truth, used for the metric geometric evaluation below;
\texttt{medium} ($100$) are real photos with a single dominant layout;
\texttt{complicated} ($100$) are real photos with cluttered, multi-object
layouts. Room types are diverse and imbalanced by design, following what
naturally occurs in each source pool rather than a fixed quota: \texttt{easy}
is mostly living/dining rooms (3D-FRONT's furnished-room distribution),
\texttt{medium} is mostly bedrooms, and \texttt{complicated} spans living
rooms, bedrooms, offices, dining rooms, kitchens, bathrooms, gyms, and a hallway.

\subsection{Per-scene contents} Each \texttt{\textless difficulty\textgreater
/\textless room\_type\textgreater\_NN/} folder contains the reference photo, the physically-lit render,
the reconstructed scene, shell + wall-mounted + furniture + ceiling objects + decorations), and the calibrated camera used to render it from Section \Cref{subsec:3d-room-layout}. Every scene additionally ships two structured annotation files: a segmentation colormap
merges the per-object masks segmented across every placement stage
(furniture, wall-mounted, ceiling, decoration) into a single image, each
object rendered in its own solid RGB colour, while the companion json records, per object, that color alongside its stage, type, source phrase, and originating mask file; and another json giving each object's size, world
position, wall affinity, and inter-object spatial relations (facing,
in-front-of, on-top-of, grouped-with) recovered during placement.

\begin{figure*}[t]
  \centering
  \includegraphics[width=\textwidth]{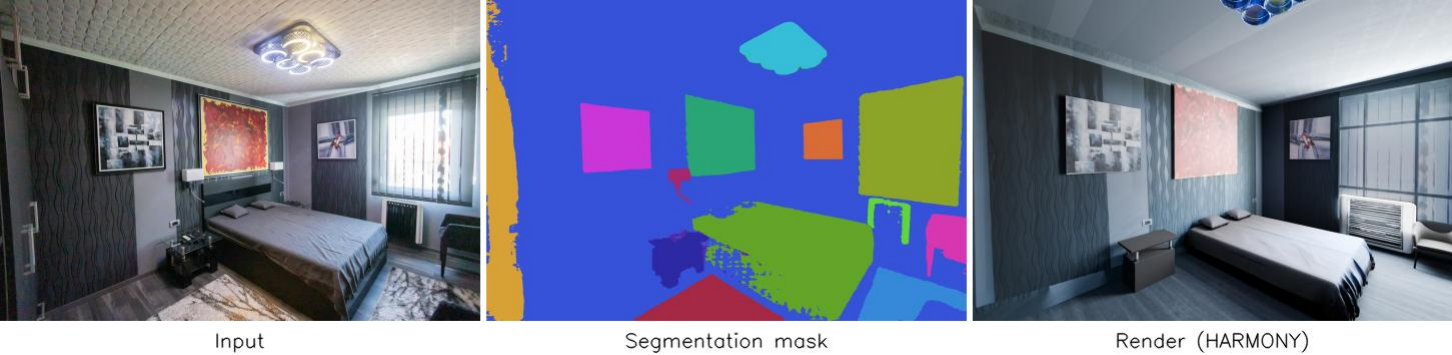}
\caption{HARMONY300 example: reference
photo, instance segmentation colormap, and \method's physically-lit render.}
\label{fig:harmony300_example}
\end{figure*}

\subsection{Sourcing and licensing} $196$ scenes are sourced from Pexels and
$2$ from Unsplash (both permissive, no attribution required), $100$ from
3D-FRONT, and $2$ synthetic renders from 3D-FUTURE (via SceneGen); the latter
two require research-only use and citation. Per-scene source URLs are listed
in the csv file included in the dataset. The dataset is released under CC-BY-NC-4.0.

\section{User Study}
\label{sec:userstudy}
A self-contained static web page in \Cref{fig:user-study-ui} shows, per scene, the input photo beside the scene's six reconstructions (\method\ + five baselines) as anonymized panels A--F; we use the $20$ scenes ($10$ front3D cases from \texttt{easy} and $10$ reallife cases from the \texttt{medium} and \texttt{hard} randomly-selected from HARMONY300 dataset) with a successful render from every method, so each trial is a complete six-way comparison. Participants assign each panel a unique rank
from $1$ (best match to the input) to $6$ (worst) so that the interface forbids
ties. The letter--method mapping is reshuffled per (participant, scene) via a
seed hashed from participant ID and scene name, so no participant sees a
consistent panel--identity association; identities are decoded only
afterward, from the stored seed. Responses autosave locally and submit to a
shared log on completion. The $16$ participants were
uncompensated volunteers, each ranking all $20$ scenes. The interface instructions read
verbatim:


\begin{figure*}[t]
  \centering
  \includegraphics[width=\textwidth]{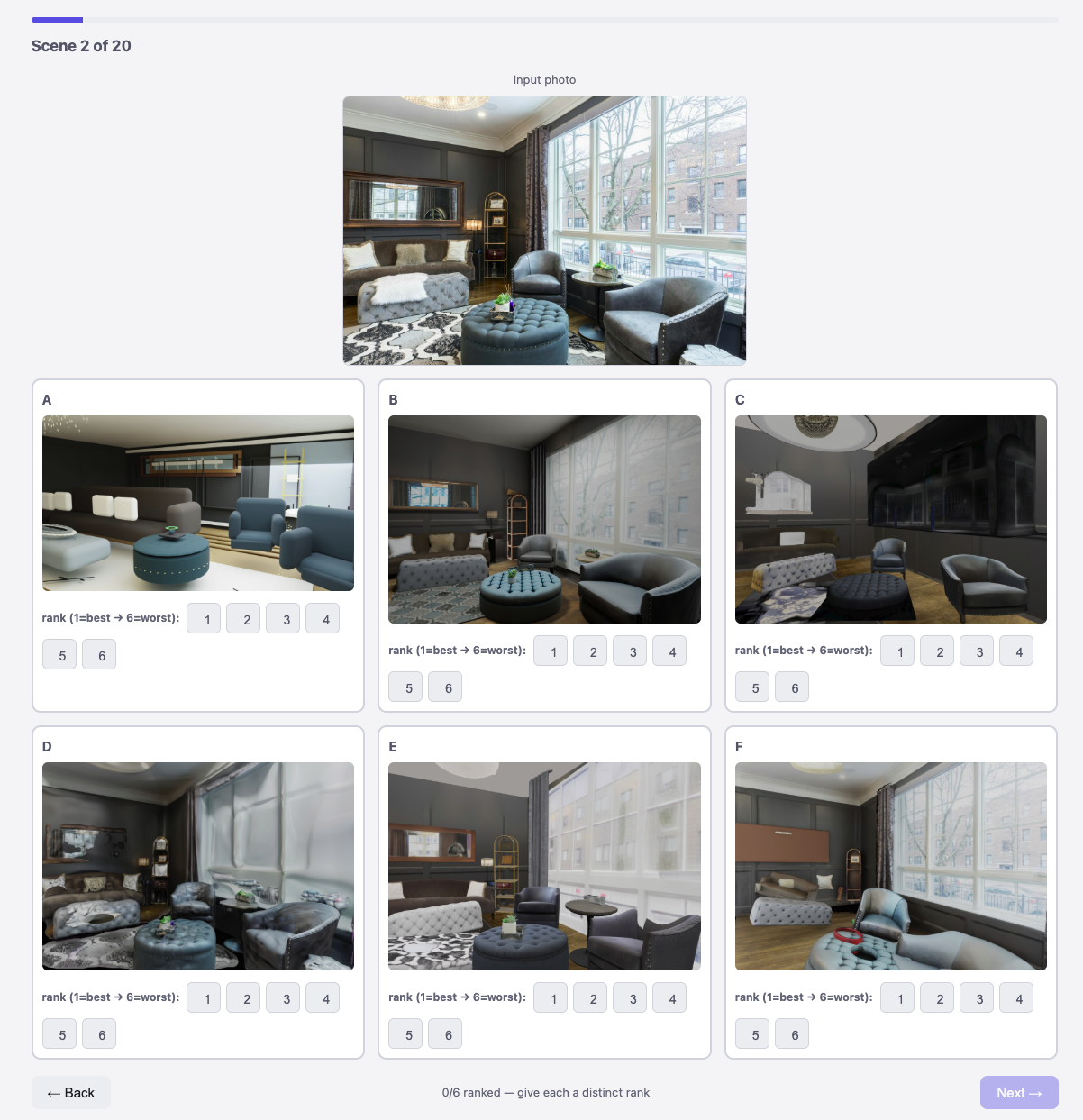}
\caption{An example study trial: the input
photo shown to participants (top) and the six candidate reconstructions
they rank (bottom, 2$\times$3) including \method\ and five baselines. In the live
interface each panel appears unlabeled as an anonymized letter A--F,
reshuffled per (participant, scene); method identities are shown here only
for illustration.}
\label{fig:user-study-ui}
\end{figure*}

\begin{tcolorbox}[
    breakable, colback=gray!10, colframe=gray!80, fontupper=\ttfamily,
    boxrule=0.5pt, arc=2mm, left=2mm, right=2mm, top=1mm, bottom=1mm
]
\par\noindent{\small\ttfamily
\par\smallskip
You'll see 20 real room photos. For each, six reconstructions (A to F, shown in
random order) attempt to recreate the scene. Rank all six by how well they
match the input photo perceptually (geometry, objects, layout, realism): give each a
distinct rank from 1 = best to 6 = worst. Clicking a rank that's already used moves it to the current one, so every rank ends up used exactly once.
\par}
\end{tcolorbox}

\section{Additional Results}
\label{sec:qual}
This section presents additional qualitative results across all three HARMONY300 splits, an alternative viewpoint for inspecting the reconstructed geometry and generalization to multi-room layouts.

\subsection{3D-Front Results}
\begin{figure*}[t]
\centering
\includegraphics[width=\textwidth]{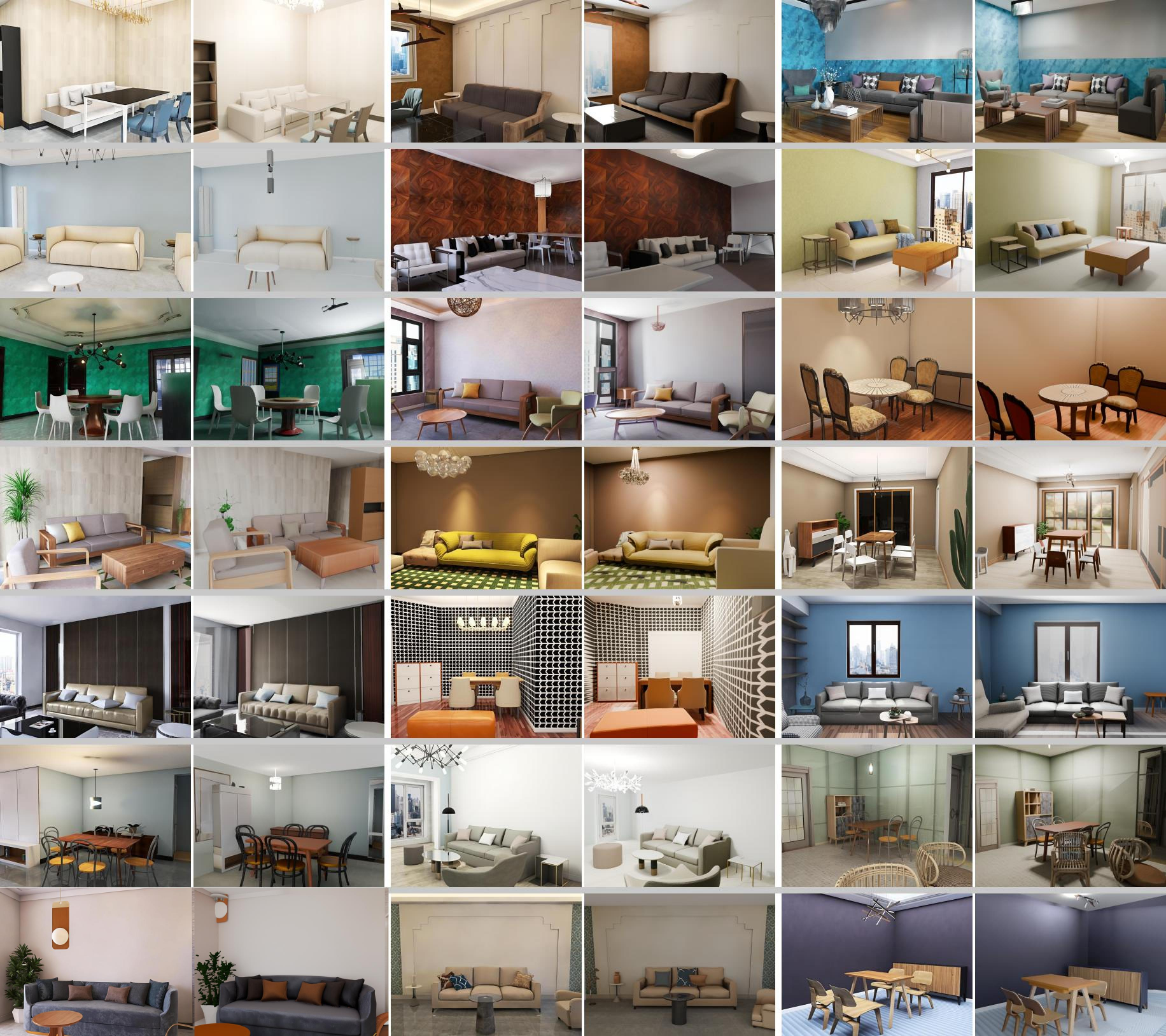}
\caption{Scenes from HARMONY300's \texttt{easy} split. For each pair: input photo (left) and \method's render (right)}
\label{fig:easy100}
\end{figure*}

\Cref{fig:easy100} shows further scenes from the synthetic \texttt{easy} split (3D-Front renders), illustrating reconstruction quality across diverse layouts, furniture styles, wall colours, and lighting conditions.

\subsection{Diverse Room Types}
\begin{figure*}[t]
  \centering
  \includegraphics[width=0.95\textwidth]{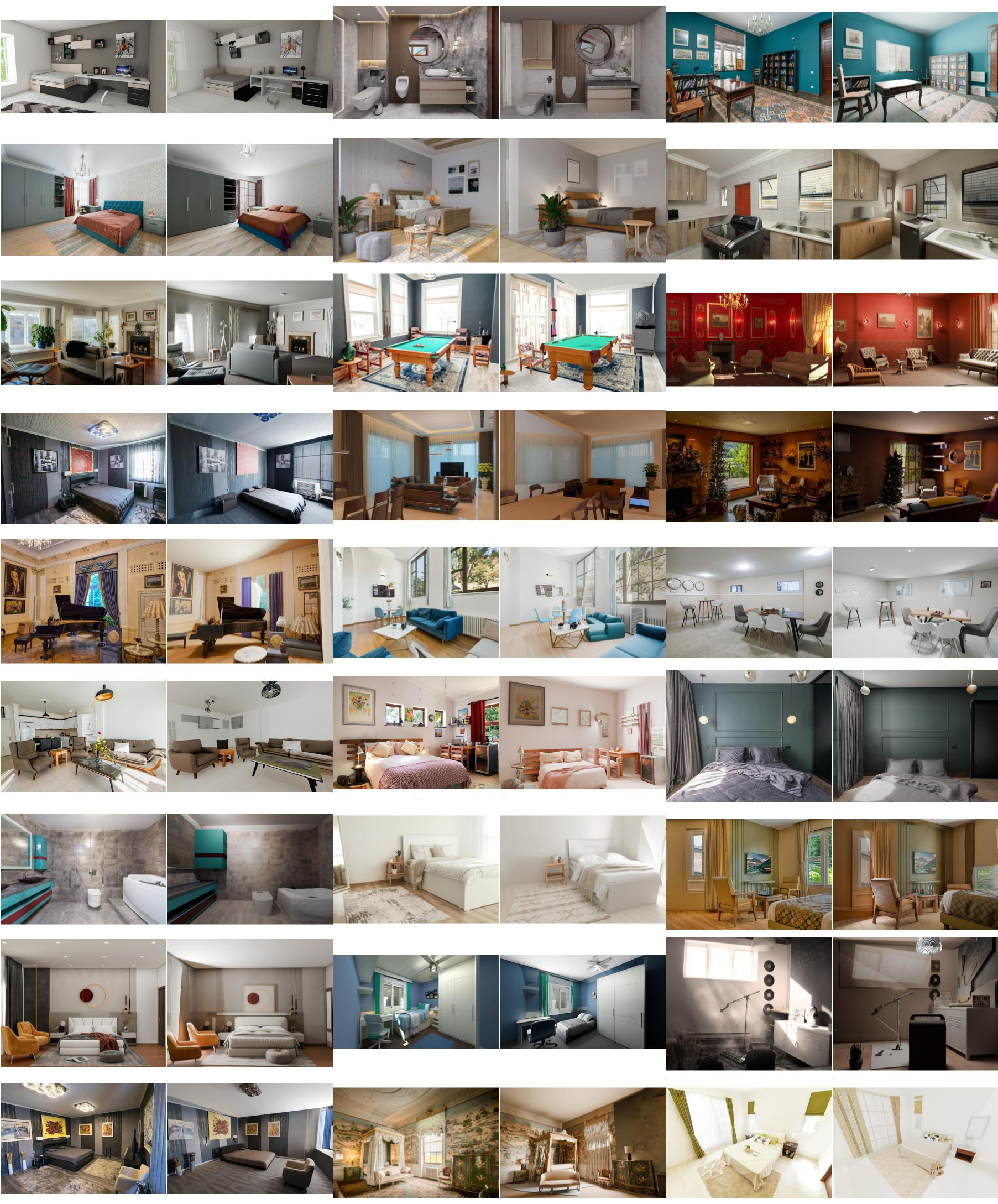}
\caption{Scenes from HARMONY300's \texttt{medium} and \texttt{complicated} splits, spanning diverse room types, layouts and lighting conditions. For each pair: input photo (left) and \method's render (right).}
\label{fig:gallery}
\end{figure*}

\Cref{fig:gallery} shows further scenes from the harder \texttt{medium} and
\texttt{complicated} splits, such as real photos spanning bedrooms, living rooms,
offices, kitchens, and bathrooms, demonstrating that reconstruction
quality holds up on cluttered, real-world layouts and not just the
synthetic \texttt{easy} split.

\subsection{Canonical Views}
\begin{figure*}[t]
  \centering
  \includegraphics[width=0.95\textwidth]{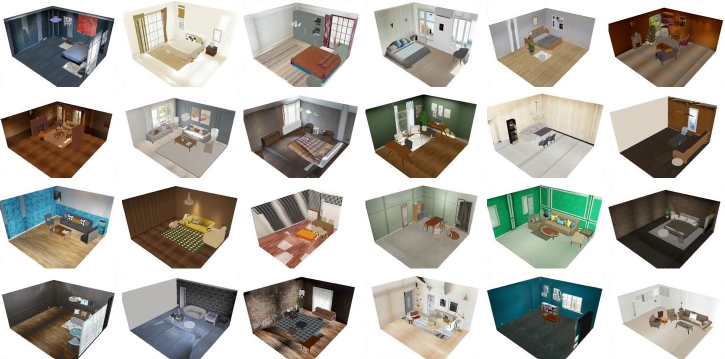}
  \caption{Reconstructed HARMONY300 scenes rendered from a canonical elevated
  viewpoint rather than the input camera, with the ceiling and near wall removed
  so the full room layout is visible.}
  \label{fig:gallery-canonical}
\end{figure*}

The galleries above render each scene from its VGGT-calibrated input camera,
which shows only the portion of the room the reference photo happened to frame.
\Cref{fig:gallery-canonical} instead renders each scene from a canonical
viewpoint in 3D, exposing the full room layout and making furniture arrangement
and geometry directly comparable across scenes, independent of how each
photograph was composed.

\subsection{Multi-Room Layout}
\begin{figure*}[t]
  \centering
  \includegraphics[width=1\textwidth]{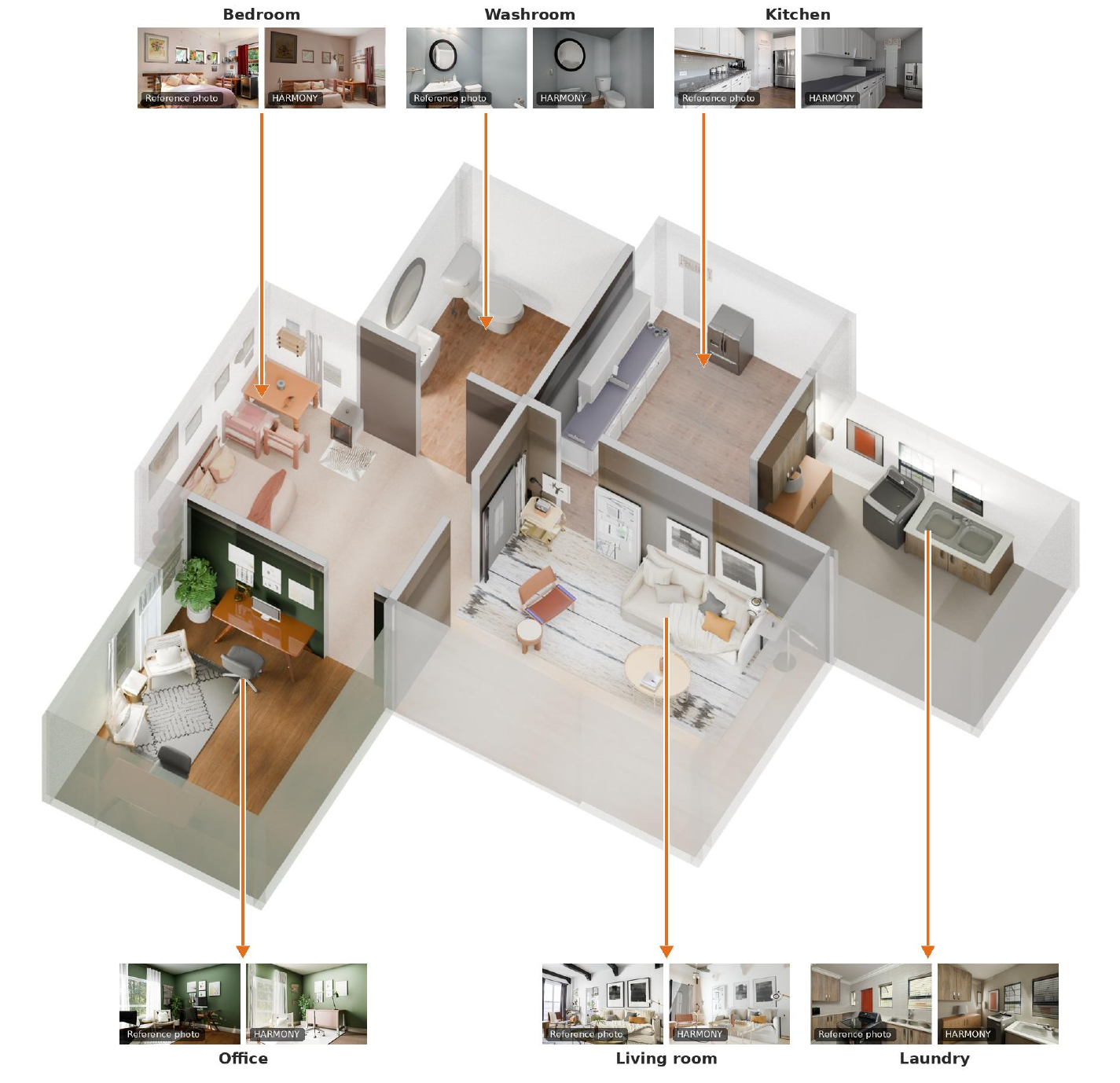}
\caption{\textbf{Multi-room layout generalization.} HARMONY adapts the reconstructed scene to diverse room configurations by refining camera poses with respect to a canonical layout.}
\label{fig:multi-room}
\end{figure*}

By refining camera poses with respect to a canonical layout, HARMONY naturally generalizes to diverse room configurations and multi-room environments. As shown in \Cref{fig:multi-room}, our framework can adapt the reconstructed scene to different layout settings while maintaining consistent spatial relationships.

\subsection{Applications}
Reconstruction quality is ultimately judged by what the scene supports downstream. We demonstrate two uses that stress different properties of the output: passive physical plausibility, and contact-rich interaction by an
embodied agent.

\subsubsection{Physical Simulations}
\begin{figure*}[t]
  \centering
  \includegraphics[width=\textwidth]{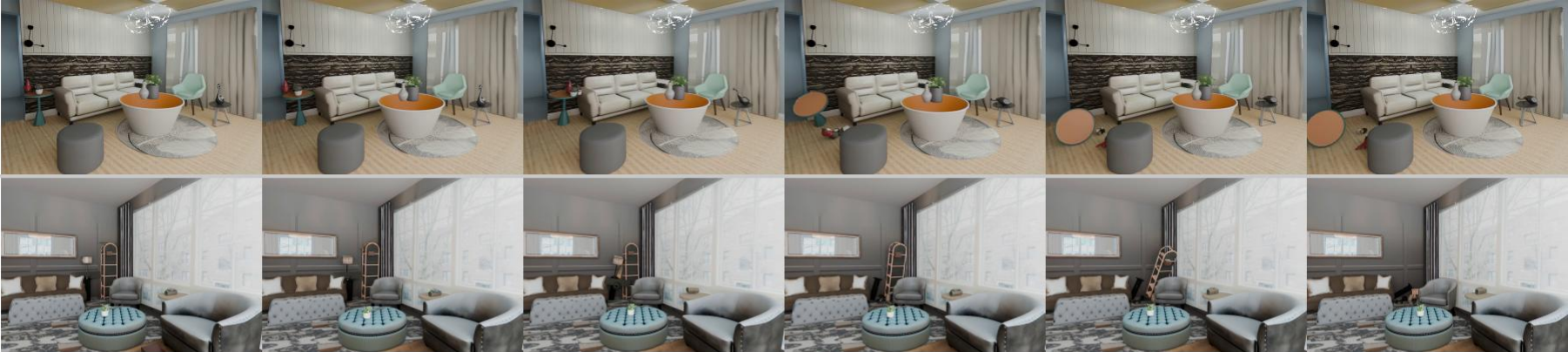}
\caption{Six evenly-spaced frames from rigid-body animations simulating earthquakes.
}
\label{fig:phys_sim}
\end{figure*}

As a downstream test that the reconstruction is physically usable, not just visually plausible, we animate each finished scene as a rigid-body
simulation, path-traced with the same Cycles + lighting as the still
render, under earthquake regime. Furniture and decorations are
genuine dynamic bodies, moving only from gravity and friction with each
object's own VLM-estimated mass and friction, so a tall bookcase topples
while a heavy sofa barely shifts. \Cref{fig:phys_sim} shows evenly-sampled frames from both regimes.

\begin{tcolorbox}[
    breakable,
    colback=gray!10,    
    colframe=gray!80,   
    fontupper=\ttfamily, 
    boxrule=0.5pt,      
    arc=2mm,            
    left=2mm,           
    right=2mm,          
    top=1mm,            
    bottom=1mm          
]
\par\noindent{\small\ttfamily
\par\smallskip
You estimate physical properties for a rigid-body simulation of this room.
For EACH item label below, give a realistic real-world mass in kilograms and
a coefficient of friction against a wood floor (0.2 = very slippery, 0.9 =
high grip). Base it on a typical reala li example of that object.
Labels: \{labels\}\par\smallskip
OUTPUT: \{"items": \{"\textless label\textgreater": \{"mass\_kg", "friction"\},
...\}\}
\par}
\end{tcolorbox}

\subsubsection{Robot Interaction}
\begin{figure*}[t]
  \centering
  \includegraphics[width=\textwidth]{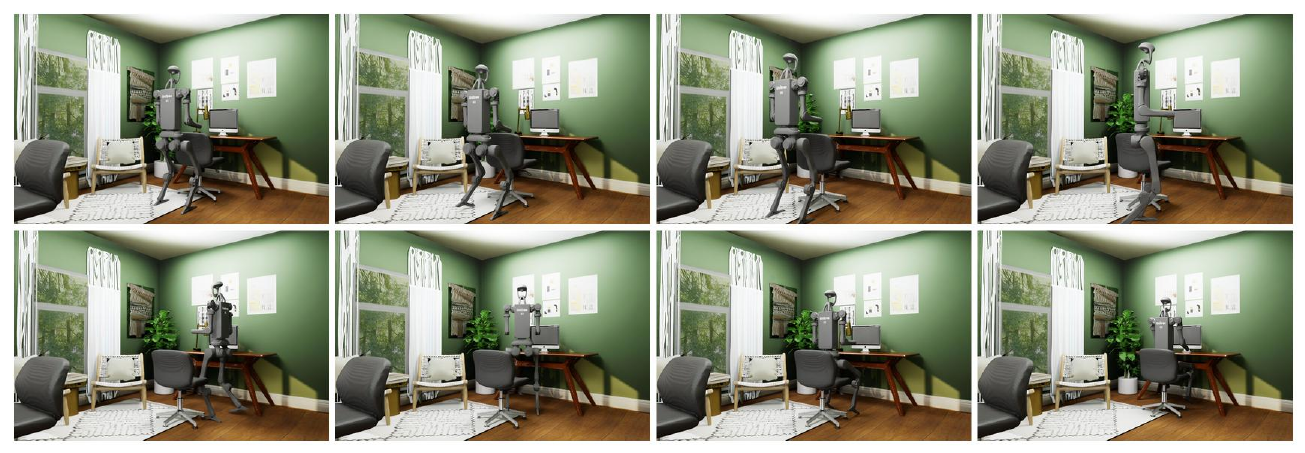}
  \caption{\textbf{Humanoid interaction in a reconstructed room.} A Unitree H1 (19 DoF) operating in a scene reconstructed
  from a single photograph by HARMONY. Ordered left to right, top to bottom: the robot
  crouches to reach the chair backrest, draws the chair away from the
  desk, releases and stands, steps around the chair, crosses to its front,
  turns, seats itself, and places both hands on the desk. All contacts are
  resolved against the reconstructed geometry: hands meet the backrest at, thighs rest on the seat pan, and both feet remain on the floor across all 230 frames.}
  \label{fig:h1_sequence}
\end{figure*}

\begin{figure}[t]
  \centering
  \includegraphics[width=\columnwidth]{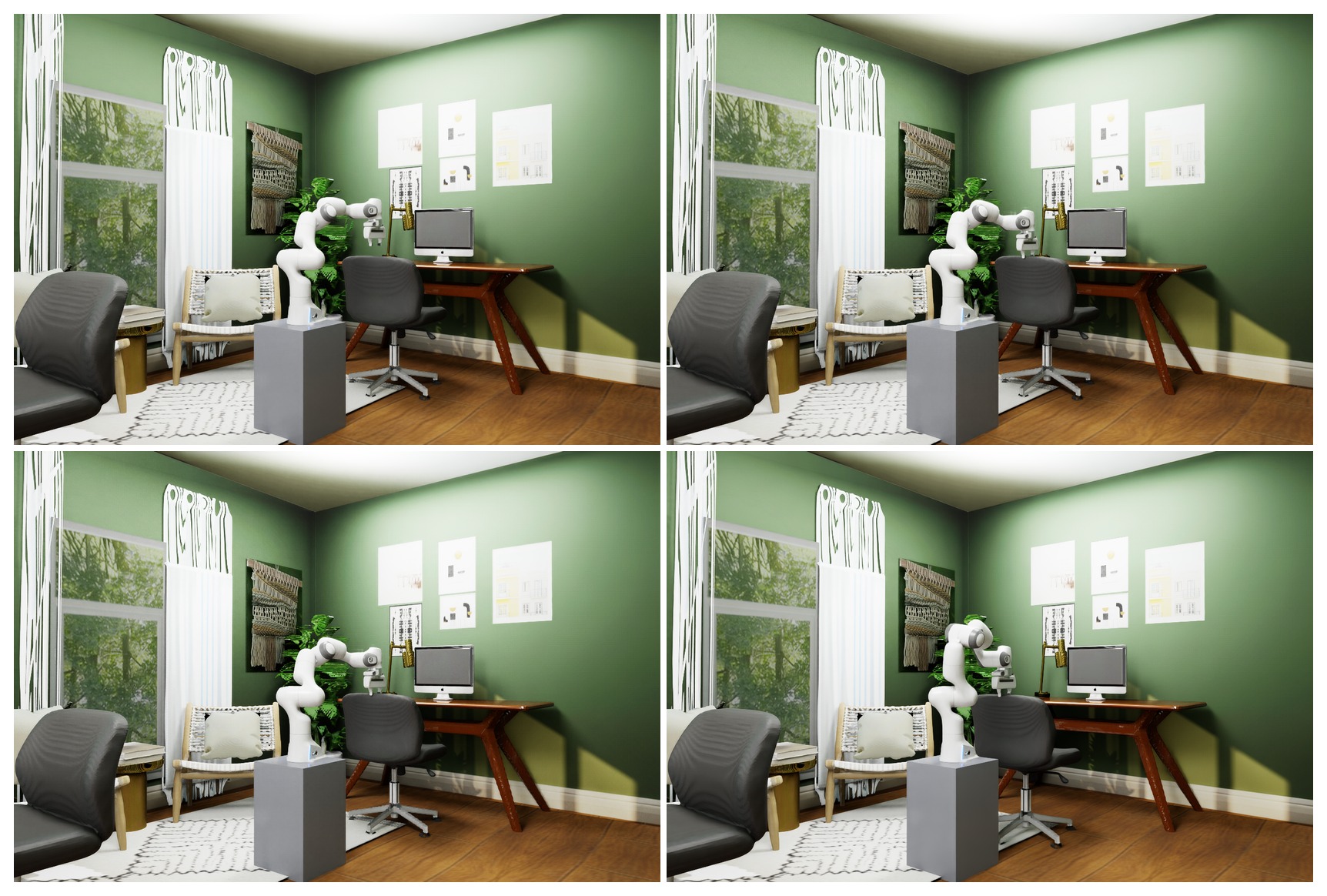}
  \caption{\textbf{Manipulator grasp of a reconstructed object.} A Franka Panda descends onto the office chair's backrest, aligns its fingers to straddle the panel, closes the jaw, and draws the chair backward. The grasp height is set from the measured panel cross-section:
  at 0.868\,m the backrest is 68\,mm thick, within the 80\,mm jaw, whereas
  35\,mm lower it thickens to 87\,mm and cannot be grasped. End-effector
  poses are solved with Lula IK (no failures across 140 frames), and the
  fingertips hold 0.868\,m throughout the 0.45\,m pull.}
  \label{fig:arm_sequence}
\end{figure}

Beyond passive dynamics, the reconstructed room is also useful as a robotics environment if its surfaces support contact-rich interaction. We import each
finished scene into Isaac Sim and drive two embodied agents through
manipulation sequences: a Unitree H1 humanoid that relocates the office chair, seats itself, and reaches the workstation, and a Franka Panda that grasps the chair backrest and draws it back. Contact targets are taken from the reconstructed geometry itself, so the interaction is grounded in the
reconstruction rather than in hand-placed proxies. Robot motion is kinematically scripted; the sequences test the geometry's affordances, not a control policy. \Cref{fig:h1_sequence,fig:arm_sequence} show sampled frames.

